herman
CAPPELEN

josh
DEVER

# MAKING AI
# INTELLIGIBLE

*philosophical foundations*

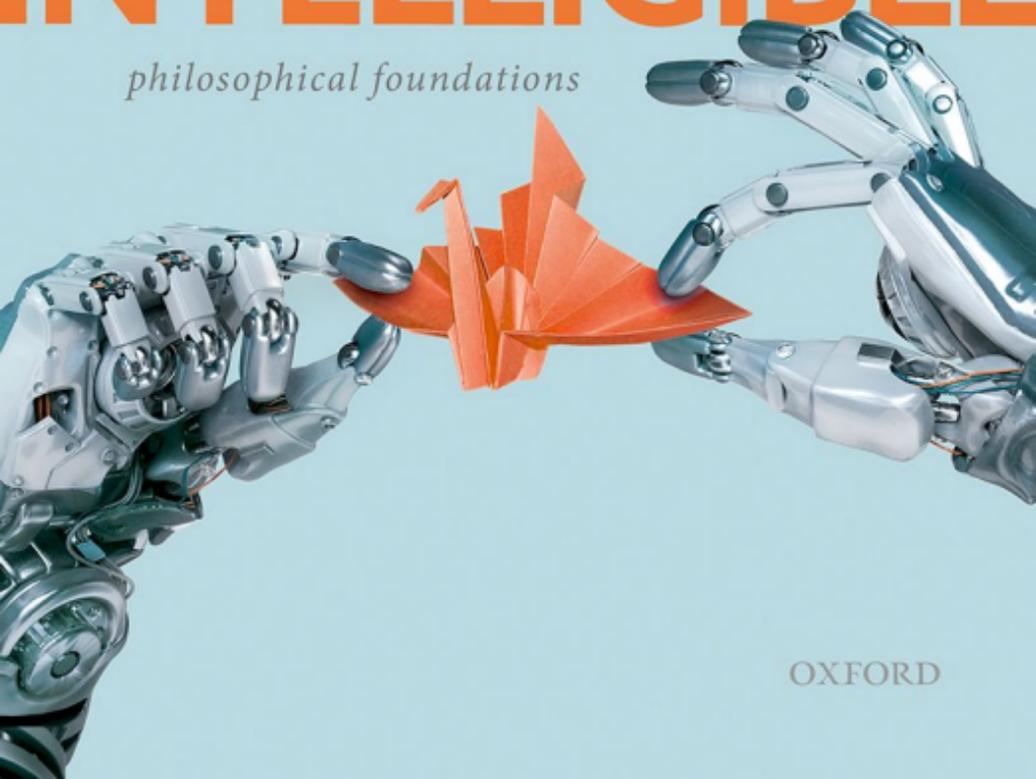

MAKING AI INTELLIGIBLE

# HERMAN CAPPELEN
# AND JOSH DEVER

# MAKING AI
# INTELLIGIBLE

## Philosophical Foundations

# OXFORD
### UNIVERSITY PRESS

# CONTENTS

## PART III: CONCLUSION

# PART I

# INTRODUCTION AND OVERVIEW

# 1

# INTRODUCTION

## The Goals of This Book: The Role of Philosophy in AI Research

This is a book about some aspects of the philosophical foundations of Artificial Intelligence. Philosophy is relevant to many aspects of AI and we don't mean to cover all of them.[1] Our focus is on one relatively underexplored question: Can philosophical theories of meaning, language, and content help us understand, explain, and maybe also improve AI systems? Our answer is 'Yes'. To show this, we first articulate some pressing issues about how to interpret and explain the outputs we get

[1] Thus we are not going to talk about the consequences that the new wave in AI might have for the empiricism/rationalism debate (see Buckner 2018), nor are we going to consider—much—the question of whether it is reasonable to say that what these programs do is 'learning' in anything like the sense with which we are familiar (Buckner 2019, 4.2), and we'll pass over interesting questions about what we can learn about philosophy of mind from deep learning (López-Rubio 2018). We are not going to talk about the clearly very important ethical issues involved, either the recondite ones, science-fictional ones (such as the paperclip maximizer and Roko's Basilisk (see e.g. Bostrom 2014 for some of these issues)), or the more down-to-earth issues about, for example, self-driving cars (Nyholm and Smids 2016, Lin et al. 2017), or racist and sexist bias in AI resulting from racist and sexist data sets (Zou and Schiebinger 2018). We also won't consider political consequences and implications for policy making (Floridi et al. 2018).

from advanced AI systems. We then use philosophical theories to answer questions like the above.

## An Illustration: Lucie's Mortgage Application is Rejected

Here is a brief story to illustrate how we use certain forms of artificial intelligence and how those uses raise pressing philosophical questions:

> Lucie needs a mortgage to buy a new house. She logs onto her bank's webpage, fills in a great deal of information about herself and her financial history, and also provides account names and passwords for all of her social media accounts. She submits this to the bank. In so doing, she gives the bank permission to access her credit score. Within a few minutes, she gets a message from her bank saying that her application has been declined. It has been declined because Lucie's credit score is too low; it's 550, which is considered very poor. No human beings were directly involved in this decision. The calculation of Lucie's credit score was done by a very sophisticated form of artificial intelligence, called SmartCredit. A natural way to put it is that this AI system *says* that Lucie has a low credit score and on that basis, another part of the AI system *decides* that Lucie should not get a mortgage.

It's natural for Lucie to wonder where this number 550 came from. This is Lucie's first question:

**Lucie's First Question**. What does the output '550' that has been assigned to me *mean*?

The bank has a ready answer to that question: the number 550 is a credit score, which represents how credit-worthy Lucie is. (Not very, unfortunately.) But being told this doesn't satisfy Lucie's unease. On reflection, what she really wants to know is *why* the output means that. This is Lucie's second question:

> **Lucie's Second Question**: Why is the '550' that the computer displays on the screen an assessment of my credit-worthiness? What *makes* it mean that?

It's then natural for Lucie to suspect that answering this question requires understanding how SmartCredit works. What's going on under the hood that led to the number 550 being assigned to Lucie? The full story gets rather technical, but the central details can be set out briefly:

*Simple Sketch of How a Neural Network Works*[2]

SmartCredit didn't begin life as a credit scoring program. Rather, it started life as a general neural network. Its building blocks are small 'neuron' programs. Each neuron is designed to take a list of input data points and apply some mathematical function to that list to produce a new output list. Different neurons can apply different functions, and even a single neuron can change, over time, which function it applies.

The neurons are then arranged into a network. That means that various neurons are interconnected, so that the output of one neuron provides part of the input to another neuron. In particular, the neurons are arranged into layers. There is a top layer of neurons—none of these neurons are connected to each other, and all of them are designed to receive input from some outside data source. Then there is a second layer. Neurons on the top layer are connected to neurons on the second layer, so that top layer neurons provide inputs to second layer neurons. Each top layer neuron is connected to every second layer neuron, but the connections also have variable weight. Suppose the top layer neurons $T_1$ and $T_2$ are connected to second layer neurons $S_1$ and $S_2$, but that the $T_1$-to-$S_1$ connection and the $T_2$-to-$S_2$ connections are weighted heavily while the $T_1$-to-$S_2$ connection and the $T_2$-to-$S_1$ connections are weighted lightly. Then the input to $S_1$ will be a mixture of the $T_1$ and $T_2$ outputs with the $T_1$ output dominating, while the input to $S_2$ will be a mixture of the $T_1$ and $T_2$ outputs with the $T_2$ output dominating. And just as the mathematical function applied by a given neuron can change, so can the weighting of connections between neurons.

---

[2]  For a gentle and quick introduction to the computer science behind basic neural networks, see Rashid 2016. A relatively demanding article-length introduction is LeCun et al. 2015, and a canonical textbook that doesn't shirk detail and is freely available online is Goodfellow et al. 2016.

After the second layer there is a third layer, and then a fourth, and so on. Eventually there is a bottom layer, the output of which is the final output of SmartCredit. The bottom layer of neurons is designed so that that final output is always some number between 1 and 1000.

The bank offers to show Lucie a diagram of the SmartCredit neural network. It's a complicated diagram—there are 10 levels, each containing 128 neurons. That means there are about 150,000 connections between neurons, each one labelled with some weight. And each neuron is marked with its particular mathematical transformation function, represented by a list of thousands of coefficients determining a particular linear transformation on a thousands-of-dimensions vector.

Lucie finds all of this rather unilluminating. She wonders what any of these complicated mathematical calculations has to do with why she can't get a loan for a new house. The bank continues explaining. So far, Lucie is told, none of this information about the neural network structure of SmartCredit explains why it's evaluating Lucie's creditworthiness. To learn about that, we need to consider the neural network's training history.

*A bit more about how SmartCredit was created*

Once the initial neural network was programmed, designers started training it. They trained it by giving it inputs of the sort that Lucie has also helpfully provided. Inputs were thus very long lists of data including demographic information (age, sex, race, residential location, and so on), financial information (bank account balances, annual income, stock holdings, income tax report contents, and so on), and an enormous body of social media data (posts liked, groups belonged to, Twitter accounts followed, and so on). In the end, all of this data is just represented as a long list of numbers. These inputs are given to the initial neural network, and some final output is produced. The programmers then evaluate that output, and give the program a score based on how acceptable its output was that measures the program's error score. If the output was a good output, the score is a low score; if the output was bad, the score is a high score. The program then responds to the score by trying to redesign its neural network to produce a lower score for the same input. There are a number of complicated mathematical methods that can be used to do the redesigning, but they all come down to making small changes in weighting and checking to see whether those small changes would have made the score lower or higher. Typically, this then means that a bunch of differential equations need to be solved. With the necessary computations done, the program adjusts its weights, and then it's ready for the next round of training.

Lucie, of course, is curious about where this scoring method came from—how do the programmers decide whether SmartCredit has done a good job in assigning a final output to input data?

*The Scoring Method*

The bank explains that the programmers started with a database of millions of old credit cases. Each case was a full demographic, financial, and social media history of a particular person, as well as a credit score that an old-fashioned human credit assessor had assigned to that person. SmartCredit was then trained on that data

set—over and over it was given inputs (case histories) from the data set, and its neural network output was scored against the original credit assessment. And over and over SmartCredit reweighted its own neural network trying to get its outputs more and more in line with the original credit assessments.

That's why, the bank explains, SmartCredit has the particular collections of weights and functions that it does in its neural network. With a different training set, the same underlying program could have developed different weights and ended up as a program for evaluating political affiliation, or for determining people's favourite movies, or just about anything that might reasonably be extracted from the mess of input social media data.

Lucie, though, finds all of this a bit too abstract to be very helpful. What she wants to know is why *she*, in particular, was assigned a score of *550*, in particular. None of this information about the neural architecture or the training history of SmartCredit seems to answer that question.

### How all this applies to Lucie

Wanting to be helpful, the bank offers to let Lucie watch the computational details of SmartCredit's assessment of Lucie's case. First they show Lucie what the input data for her case looks like. It's a list of about 100,000 integers. The bank can tell Lucie a bit about the meaning of that list—they explain that one number represents the number of Twitter followers she has, and another number represents the number of times she has 'liked' commercial postings on Facebook, and so on.

Then they show Lucie how that initial data is processed by SmartCredit. Here things become more obscure. Lucie can watch the computations filter their way down the neural network. Each neuron receives an input list and produces an output list, and those output lists are combined using network weightings to produce inputs for subsequent neurons. Eventually, sure enough, the number '550' drops out of the bottom layer.

But Lucie feels rather unilluminated by that cascading sequence of numbers. She points to one neuron in the middle of the network and to the first number (13,483) in the output sequence of that neuron. What, she asks, does that particular number mean? What is *it* saying about Lucie's credit worthiness? This is Lucie's third question:

> **Lucie's Third Question**: How is the final meaningful state of SmartCredit (the output '550', meaning that Lucie's credit score is 550) the result of other meaningful considerations that SmartCredit is taking into account?

The bank initially insists that that question doesn't really have an answer. That particular neuron's output doesn't by itself mean anything—it's just part of a big computational procedure that holistically yields an assessment of Lucie's credit worthiness. No particular point in the network can be said to mean anything in particular—it's the network as a whole that's telling the bank something.

Lucie is understandably somewhat sceptical at this point. How, she wonders, can a bunch of mathematical transformations, none of which in particular can be tied to any meaningful assessment of her credit-worthiness, somehow all add up to saying something about whether she should get a loan? So she tries a different approach. Maybe looking at the low-level computational details of SmartCredit isn't going to be illuminating, but perhaps she can at least be told what it was in her history that SmartCredit found objectionable. Was it her low annual income that was responsible? Was it those late credit card payments in her early twenties? Or was it the fact that she follows a number of fans of French film on Twitter? Lucie here is trying her third question again—she is still looking for other meaningful states of SmartCredit that explain its final meaningful output, but no longer insisting that those meaningful states be tied to specific low-level neuron conditions of the program.

Unfortunately, the bank doesn't have much helpful to say about this, either. It's easy enough to spot particular variables in the initial data set—the bank can show her where in the input her annual income is, and where her credit card payment history is, and where her Twitter follows are. But they don't have much to say about how SmartCredit then assesses these different factors. All they can do is point again to the cascading sequence of calculations—there are the initial numbers, and then there are millions upon millions of mathematical operations on those initial numbers, eventually dropping out a final output number. The bank explains that that huge sequence of mathematical operations is just too long and complicated to be humanly understood—there's just no point in trying to follow the details of what's going on. No one could hold all of those numbers in their head, and even if they could, it's not clear that doing so would lead to any real insight into what features of the case led to the final credit score.

## Abstraction: The Relevant Features of the Systems We Will be Concerned with in This Book

Our concern is not with any particular algorithm or AI systems. It is also not with any particular way of creating a neural network. These will change over time and the cutting edge of programming today will seem dated in just a year or two. To identify what we will be concerned with, we must first distinguish two levels at which an AI system can be characterized:

- On the one hand, it is an abstract mathematical structure. As such it exists outside space and time (it is not located anywhere, has no weight, and doesn't start existing at any particular point in time).
- However, when humans use and engage with AI, they have to engage with something that exists as a physical object, something they can see or hear or feel. This will be the **physical implementation** (or **realization**) of the abstract structure. When Lucie's application was rejected, the rejection was presented to her as a token of numbers and letters on a computer screen. These were physical phenomena, generated by silicon chips, various kinds of wires, and other physical things (many of them in different locations around the world).

This book is not about a particular set of silicon chips and wires. It is also not about any particular program construed as an abstract object. So we owe you an account of what the book is about. Here is a partial characterization of what we have in mind when we talk about 'the outputs of AI systems' in what follows:[3]

- The output (e.g. the token of '550' that occurs on a particular screen) is produced by things that are not human. The non-human status of the producer can matter in at least three ways:

  First, these programs don't have the same kind of physical imple­mentation as our brains do. They may use 'neurons', but their neurons are not the same kind of things as our neurons—they differ of course physically (being non-biological), but also computationally (they don't process inputs and produce outputs in the same way as our neurons). And their neurons are massively different in number and arrangement from our neurons, and massively different in the way they dynamically respond to feedback.

---

[3] This is not an effort to specify necessary and sufficient conditions for being an AI system—that's not a project we think is productive or achievable.

Second, these programs don't have the same abilities as we do. We have emotional repertoires and sensory experiences they lack, and arguably have beliefs, desires, hopes, and fears that they also lack. On the other hand, they have computational speeds and accuracies that we lack.

Third, these programs don't have the same histories that we do. They haven't had the kind of childhoods we have had, and in particular haven't undergone the same experiences of language acquisition and learning that we have. In short, they are non-human (where we will leave the precise characterization of this somewhat vague and open-ended).

- When we look under the hood—as Lucie did in the story above—what we find is not intelligible to us. It's a black box. It will operate in ways that are too complex for us to understand. It's important to highlight right away that this particular feature doesn't distinguish it from humans: when you look under the hood of a human, what you will find is brain tissue—and at a higher level, what looks like an immensely complex neutral network. In that sense, the human mind is also a black box, but as we pointed out above, the physical material under the hood/skull is radically different.

- The systems we are concerned with are made by human programmers with their own beliefs and plans. As Lucie saw, understanding SmartCredit requires looking beyond the program itself to the way that the program was trained. But the training was done by people, who selected an initial range of data, assigned target scores to those initial training cases based on their own plans for what the program should track, and created specific dynamic methods for the program to adjust its neural network in the face of training feedback.

- The systems we are concerned with are systems that are intended to play a specific role, and are perceived as playing that role. SmartCredit isn't just some 'found artefact' that's a mysterious black box for transforming some numbers into other numbers. It's a program that occupies a specific social role: it was designed specifically to assign credit scores, and it's used by banks because it's perceived as assigning credit scores. It's treated as useful, as producing outputs that really are meaningful and helpful credit scores, and it becomes entrenched in the social role it occupies because it's perceived as useful in that way.

None of this adds up to a complete metaphysics of AI systems. That's not the aim of this book. Instead, we hope it puts readers in a position to identify at least a large range of core cases.

## The Ubiquity of AI Decision-Making

SmartCredit raises concerns about what its outputs mean. But SmartCredit is only the tip of the iceberg. We are increasingly surrounded by AI systems that use neural network machine learning methods to perform various sorts of classifications. Image recognition software classifies faces for security purposes, tags photographs on social media, performs handwriting analysis, guides military drones to their targets, and identifies obstacles and street signs for self-driving cars. But AI systems of this sort aren't limited to simple classification tasks. The same underlying neural network programming methods give rise, for example, to strategic game-playing. Google's AlphaZero has famously achieved superhuman levels of performance in chess, Go, and Shogi. Other machine learning approaches have been applied to a wide variety of games, including video games such as Pac-Man, Doom, and Minecraft.[4] Other AI systems perform variants of the kind of 'expert system' recommendation as SmartCredit. Already there are AI systems that attempt to categorize skin lesions as cancerous or not, separate spam emails and malware from useful emails, determine whether building permits should be granted and whether prisoners should receive parole, figure out whether children are being naughty or nice using video surveillance, and work out people's sexual orientations from photographs of their faces. Other AI systems use machine learning to make predictions. For example, product recommendation software attempts to extrapolate from earlier purchases to likely future purchases, and traffic software attempts to predict future locations of congestion based on earlier traffic conditions. Machine learning can also be used for data mining, in which large quantities of data are analysed to try to find new and unexpected patterns. For example, the data mining program Word2Vec extracted from a database of old scientific papers new and unexpected scientific conclusions about thermoelectric materials.

These AI systems are able to perform certain tasks at extraordinarily high levels of precision and accuracy—identifying certain patterns much more reliably, and on the basis of much noisier input, than we can, and making certain kinds of strategic decisions with much higher accuracy than we can—and both their sophistication and their number are rapidly increasing. We should expect that in the future many of our interactions with the world will be mediated by AI systems, and many of our current intellectual activities will be replaced or augmented by AI systems.

---

[4] See https://www.sciencenews.org/article/ai-learns-playing-video-games-starcraft-minecraft for some discussion about the state and importance of AI in gaming.

Given all that, it would be nice to know what these AI systems mean. That means we want to know two things. First, we want to know what the AI systems mean with their explicit outputs. When the legal software displays the word 'guilty', does it really *mean* that the defendant is guilty? Is guilt really what the software is tracking? Second, we want to know what contentful states the AI systems have that aren't being explicitly revealed. When AlphaZero makes a chess move, is it making it for reasons that we can understand? When SmartCredit gives Lucie a credit score of 550, is it weighing certain factors and not others?

If we can't assign contents to AI systems, and we can't know what they mean, then we can't in some important sense understand our interactions with them. If Lucie is denied a loan by SmartCredit, she wants to understand why SmartCredit denied the loan. That matters to Lucie, both practically (she'd like to know what she needs to change to have a better chance at a loan next time) and morally (understanding why helps Lucie not view her treatment as capricious). And it matters to the bank and to us. If we can't tell why SmartCredit is making the decisions that it is, then we will find it much harder to figure out when and why SmartCredit is making its occasional errors.

As AI systems take on a larger and larger role in our lives, these considerations of understanding become increasingly important. We don't want to live in a world in which we are imprisoned for reasons we can't understand, subject to invasive medical conditions for reasons we can't understand, told whom to marry and when to have children for reasons we can't understand. The use of AI systems in scientific and intellectual research won't be very productive if it can only give us results without explanations (a neural network that assures us that the ABC conjecture is true without being able to tell us *why* it is true isn't much use). And things are even worse if such programs start announcing scientific results using categories that we're not sure we know the content of.

We are in danger, then, of finding ourselves living in an increasingly meaningless world. And as we've seen, it's a pressing danger, because if there is meaning to be found in the states and activities of these AI systems, it's not easily found by looking under the hood and considering their programming. Looking under the hood, all we see is jumbles of neurons passing around jumbles of numbers.

But at the same time, there's reason for optimism. After all, if you look under *our* hoods, you also see jumbles of neurons, this time passing around jumbles of electrical impulses. That hasn't gotten in the way of our producing meaningful outputs and having meaningful internal states. The hope then is that reflecting on how *we* manage to achieve meaning might help us understand how AI systems also achieve meaning.

However, we also want to emphasize that it's a guarded hope. Neural network programs are a little like us, but only a little. They are also very different in ways that will come out in our subsequent discussion. Both philosophy and science fiction have had an eye from time to time on the problem of communicating with and understanding aliens, but the aliens considered have never really been all that alien. In science fiction, we get the alien language in Star Trek's Darmok,[5] which turns out to be basically English with more of a literary flourish, the heptapod language of 'Story of Your Life',[6] which uses a two-dimensional syntax to present in a mildly encoded way what look like familiar contents, and the Quintans of Stanislaw Lem's 1986 novel *Fiasco*, who are profoundly culturally incomprehensible but whose occasional linguistic utterances have straightforward contents. In philosophy, consideration of alien languages either starts with the assumptions that the aliens share with us a basic cognitive architecture of beliefs, desires, reasons, and actions, or (as Davidson does) concludes that if the aliens aren't that much like us, then whatever they do simply can't count as a language.

---

[5]  See *Star Trek: The Next Generation*, season 5 episode 2.
[6]  In Chiang, *Stories of Your Life And Others,* Tor Books, 2002. The book was the inspiration for the film Arrival.

Our point is that the aliens are already among us, and they're much more alien than our idle contemplation of aliens would have led us to suspect. Not only that, but they are *weirdly* alien—we have built our own aliens, so they are simultaneously alien and familiar. That's an exciting philosophical opportunity—our understanding of philosophical concepts becomes deeper and richer by confronting cases that take us outside our familiar territory. We want simultaneously to explore the prospect of taking what we already know about how familiar creatures like us come to have content and using that knowledge to make progress in understanding how AI systems have content, and also see what the prospects are for learning how the notions of meaning and content might need to be broadened and expanded to deal with these new cases.

## The Central Questions of this Book

Philosophy can help us understand many aspects of AI. There are salient moral questions such as whether we *should* let AI play these important social roles. What are the moral and social consequences of letting AI systems make important decisions that throughout our history have been made by humans who could be held accountable? There are also pressing questions about whether advanced AI systems could eventually make humans superfluous—this is sometimes discussed under the label 'existential risk' of AI (see Bostrom 2014). None of these is the topic of this book.

The questions we will be concerned with have to do with **how we can interpret and understand the outputs of AI systems**. They are illustrated by the questions that Lucie asked the bank in our little story above. Recall Lucie's first question:

> **Lucie's First Question**: What does the output '550' that has been assigned to me mean?

Lucie's first question is a question about how to understand a specific output of a specific program. We're not going to try to answer Lucie's question, or even to give particular tools for answering this kind of question. But we are interested in the meta-question about whether Lucie's question is a reasonable and important one. We've already observed that AI systems are frequently used *as if* questions like Lucie's made sense and had good answers—we treat these systems *as if* they are giving us specific information about the world. It's thus important to consider whether there is a sensible way to think about these programs on which questions like Lucie's first question could eventually be answered.

This perspective leads to Lucie's second question:

> **Lucie's Second Question**: Why is the '550' that the computer displays on the screen an assessment of my credit-worthiness? What *makes* it mean that?

Our central interest in this book starts with examining what kinds of answers this question could have. If the output states of AI systems do mean something, then surely there must be some *reason* they mean what they do. If we could at least figure out what those reasons are, we might be better positioned down the road to answering Lucie's first question.

The bank tried one particular method of answering Lucie's second question: they directed Lucie to the details of SmartCredit's programming. As we saw, this method wasn't obviously successful—learning all the low-level neural network details of SmartCredit's programming didn't seem to give a lot of insight into why its outputs meant something about Lucie's credit worthiness.

But that was just one method. Our central project is to emphasize that there are many other methods that are worth considering. One way to think about the project is to remember that humans, too, are content-bearing. Our outputs, like SmartCredit's outputs, at least prima facie, mean things and carry information about the world. But looking inside our skulls for an explanation of those contents isn't likely to be much more illuminating than looking inside SmartCredit's programming code was. We emphasized above that programs like SmartCredit are different from people in many important ways, and that's worth keeping in mind (and will guide much of our discussion below). But at the same time, both we and machine-learning programs like SmartCredit are systems producing outputs based on some enormously complicated and not obviously illuminating underlying computational procedure.

That fact about *us*, though, hasn't stopped us from assigning contents to people's outputs, and it hasn't stopped us from entertaining theories about why people's outputs mean what they do. It's just forced us to consider factors other than neuro-computational implementation in answering that 'why' question. Theories about why human outputs mean what they do have appealed to mental states, to causal connections with the environment, to normative considerations of coherence and charity, to biological teleology, and to relations of social embedding. One of our central projects is then to see whether these kinds of theories can be helpfully deployed in answering Lucie's second question, and how such theories might need to be adapted to accommodate the differences between people and programs.

Lucie had a third question:

(**Lucie's Third Question**): How is the final meaningful state of SmartCredit (the output '550' meaning that Lucie's credit score is 550) the result of other meaningful considerations that SmartCredit is taking into account?

Eventually we want a good theory of content for AI systems. A good theory of content for people needs to do more than just assign contents to the things we say—it also needs to assign contents to 'hidden' internal states of beliefs and desires, which then help make sense of, and perhaps constrain the contents of, the things we say. We should be open to the possibility that it's the same for AI systems. SmartCredit 'says' some things—it produces explicit outputs of the form of the '550' evaluation it outputs for Lucie. But in making sense of why SmartCredit's explicit outputs have the meanings that they do, we might want to attribute additional contentful states to the program—for example, we might (as Lucie does) want to be able to attribute to SmartCredit various reasons that led it to assign Lucie the credit score that it did.

On a more abstract level: AI systems produce various outputs, and we can always ask what, if anything, makes it the case that an AI system has a certain output; and AI systems produce those outputs for various reasons, and we can ask whether those reasons are contentful reasons (rather than just irreducibly complicated mathematical computations), and what, if anything, makes it the case that the reasons have the contents that they do.

The underlying facts are not in dispute: ML (machine learning) systems are (or consist of) massively complex algorithms that generate an enormous neural network with thousands or millions of interconnected 'neurons'. It is also beyond dispute that in many cases the overall structure and dynamics of that system is too complex for any human to comprehend. A burning question is now: when this system produces an output consisting of an English sentence like the examples given above, how can that output mean what those English words mean? How can we know that it tells us something about what we call creditworthiness?

## 'Content? That's So 1980'

A central aim in this book is to encourage increased interaction between two groups. First, AI researchers, who are producing machine learning systems of rapidly increasing sophistication, systems that look to have the potential to take on or supplement many of our ordinary processes of reasoning, deciding, planning, and sorting. And second, philosophers, who work in a rich intellectual tradition, which provides tools for thinking about content, tools directed both at determining what features of a system make it contentful (and in what ways) and at characterizing different kinds of contents with a variety of formal tools. We want to encourage that interaction because we think that AI has a content problem—we need to be able to attribute contents to AI systems, but we're currently poorly positioned to do so.

A certain view of the history of AI research can make all of that seem like a confused retrograde step. AI researchers *tried* approaches centred around content and representation. That's what the symbolic artificial intelligence program was about, that's what led to endless projects focused on a small block world based on clear representational systems. But the wave of contemporary successes in AI has been won by *moving away* from the symbolic approaches. Neural network machine learning systems are deliberately designed *not* to start with a representational system—the whole goal is to allow data that hasn't been pre-processed into representational chunks to be filtered by neural network systems in a way that isn't mediated by representational rule systems and still produce powerful outputs. So if what we're suggesting in this book is a return to symbolic AI, and a move away from the machine learning successes, contemporary AI researchers would be understandably uninterested. (For an introduction to this sort of old-school philosophical theorizing inspired by old-school AI theorizing, see Rescorla 2015.)

But that's not what we are suggesting. Our point, in fact, is that philosophy brings to the table a collection of tools designed to find content in the wild, rather than building content into the architecture. The central problem in the philosophical study of content is this: when people go about in the world, encountering and interacting with various objects, making various sounds, having various things going on inside their heads, a bunch of contents typically result. Some of the sounds they make have contents; some of the brain states they are in have content. Philosophical accounts of content want to explain what makes that the case: what needs to be going on so that some sounds are contentful and others are not; what needs to be going on so that some sounds mean that it's raining and other sounds mean that it's sunny.

People, of course, are the original neural network systems. So the philosophical project of content must be compatible with application to neural networks. That's because the philosophical project isn't to *build* contentful systems by setting them up with the right representational tools, but rather to *understand* the contents that we find 'in the wild'. Work in philosophy of language and formal semantics has indeed produced very sophisticated mathematical models of representation. But the philosopher's suggestion isn't that we should take those models and use them in designing good humans. We (philosophers who work on the theory of content) are not proposing that babies be pre-fitted with Montague semantics, or that axiomatic theories of meaning be taught in infancy. We just want to understand the content that certain complex systems (like people) carry, whatever the causal and historical story about how they came to carry that content.

So even if the history of AI research has made you a representation/content pessimist, we encourage you to read on. We think that intellectual engagement between philosophy and AI research has the promise of letting you have your non-content-oriented design tools and your *post facto* content attribution, too. And, we want to suggest, that's a good thing, because content plays crucial roles, and AI systems that lie wholly outside the domain of content won't give us what we want.

## What This Book is **Not** About: Consciousness and Whether 'Strong AI' is Possible

There's an earlier philosophical literature on AI that we want to distance ourselves from. In influential work, John Searle (1980) distinguished between what he called Strong and Weak AI. Strong AI, according to Searle, has as a goal to create thinking agents. The aim of that research project is to create machines that *really* can think and have other cognitive states that we humans have. Searle contrasted this with the weak AI project according to which the aim was to create machines that have the *appearance* of thinking (and understanding and other cognitive states). Searle's central argument against Strong AI was the Chinese Room Argument. There's now a very big literature on the soundness of that argument (and also on how to best present the argument—for some discussion and references, see Cole 2014).

The project of this book will not engage with Searle-style arguments and we are not interested in the Strong vs Weak AI debate.

Our starting point and methodology are different from the literature in that tradition: Our goal in the first four chapters of Part I is to use contemporary theories of *semantics* and *meta-semantics* to determine whether (and how) ML systems could be interpreted. We take some of the leading theories of how language has representational properties and see what those theories have to say about ML systems. In most cases they are mixed: there's some match with what we are doing and some mismatch—and then we suggest fixes. In Chapter Four we suggest that maybe the right attitude to take is that we need to revise our meta-semantics to accommodate ML systems. Rather than use anthropocentric theories of content (i.e. theories of content based on how human language gets content) to determine whether ML systems have content, we should revise our theories of content attribution so that ML systems can be considered representational (in effect revising what representation is, so that ML systems can be accommodated).

This strategy contrasts with the argumentative strategy exemplified by Searle's Chinese Room argument (and the tradition arising from that argument): the idea behind that strategy is to use reflections on a cluster of thought experiments to settle, once and for all, the question of whether machines can understand and have a semantics. This book doesn't engage with and only indirectly takes a stand on that form of argument.

## Connection to the Explainable AI Movement

In 2018, the European Union introduced what it calls the General Data Protection Regulation. This regulation creates a 'right to explanation' and that right threatens to be incompatible with credit scores produced by neural networks (see Kaminski 2019 Goodman and Flaxman 2017, Adadi and Berrada 2018) because, as we pointed out above, many ML systems make decisions and recommendations without providing any explanation of those decisions and recommendations. Without explanations of this sort, ML systems are uninterpretable in their reasoning, and may even become uninterpretable in their results.

The burgeoning field of explainable AI (XAI) aims to create AI systems that are *interpretable* by us, that produce decisions that come with comprehensible *explanations*, that use *concepts* that we can understand, and that we can *talk to* in the way that we can engage with other rational thinkers.[7] The opacity of ML systems especially highlights the need for artificial intelligence to be both explicable and interpretable. As Ribero et al. (2016:3–4) put it, 'if hundreds or thousands of features significantly contribute to a prediction, it is not reasonable to expect any user to comprehend why the prediction was made, even if individual weights can be inspected'. But the quest for XAI is hampered both by implementation difficulties in extracting explanations of ML system behaviour and by the more fundamental problem that it is not clear what exactly explicability and interpretability *are* or what kinds of tools allow, even in-principle, achievement of interpretability.

Doshi-Velez and Kim (2017) state the goal of interpretability as being 'to explain or present [the outputs of AI systems] in understandable terms' (2017:2) and proceed to point to real problems in modelling explanations. What they do not note is that both *understanding* and *terms/concepts* require at least as much clarification as explanation. As they point out, much work in this field relies on 'know it when you see it' conceptions of these core concepts. This book aims to show how philosophy can be used to remedy this lacuna in the literature.

The core chapters in this book aim to present proposals for how we can attribute content to AI systems. We return to the implications of this for the explainable AI movement towards the end of the book.

---

[7] A recent discussion in philosophy is Páez 2019. Outside of philosophy, some recent overviews of the literature are Mueller et al. 2019 and Addadi and Berrada 2018. For particular proposals about how to implement XAI, see Ribeiro et al. 2016, Doshi-Velez & Kim 2017, and Hendricks et al. 2016. For an approach that uses some ideas from philosophy to explain 'explanation', see Miller 2018.

# Broad and Narrow Questions
## about Representation

We should note one limitation of our approach: we focus on whether the outputs of AI systems are content-bearing. In the little story about Lucie, we ask what the output '550' means. We are interested in whether that token can and should be considered contentful—in Chapter Three we put this as the question of whether there's aboutness in the AI system. This is closely connected to, but distinct from, a broader issue: Does the AI system have the ability to reason? Does it have a richer set of beliefs and so a richer set of contents? We can also ask: what is the connection between being able to represent the thought that Lucie's credit score is 550, and having a range of other thoughts about Lucie? Can a system have the ability to think only *one* thought, or does that ability by necessity come with a broader range of representational capacities? These are crucial questions that we will return to in the final chapter. Prior to that, our goal is somewhat more narrow and modest: Can we get the idea of content/representation/aboutness for AI systems off the ground at all? Are there any plausible extensions of existing meta-semantic theories that opens the door to this? Our answer is yes. In the light of that positive answer, the broader questions take prominence: how much content should be attributed? What particular content should be attributed to a particular AI system? Does SmartCredit understand 'credit worthiness', and also 'credit' and 'worthiness', and grasp the relevant compositional rule? Does understanding 'credit' involve an understanding of *money, borrowing, history*, etc? These are questions that become pressing, if the conclusions in this book are correct.

## Our Interlocutor: Alfred, The Dismissive Sceptic

A character called Alfred is central to the narrative of this book. Alfred, we imagine, is someone whose job it is to make AI systems. He is very sceptical that philosophers can contribute to his work at all. In the next chapter, Alfred is having a conversation with a philosopher. Alfred argues that while he thinks talking to philosophers is a bit interesting, it is basically useless for him. According to Alfred, philosophers have nothing substantive to contribute to the development of AI.

Alfred will return at several junctions in this book. In writing this book (and thinking through these issues), Alfred has been very useful to us—we hope he is also of some interest to readers (and especially those potential readers who are entirely unconvinced that AI will profit from an injection of philosophy).

## Who is This Book for?

The intended audience for this book are readers interested in starting to think how philosophy can help answer important questions about interpretable AI. They have some knowledge of philosophy, some knowledge of AI, and are interested in how to use the former to reflect on the latter.

There are some people who should not buy or read this book:

- If you are looking for a technical book that engages in great detail with the formal aspects of neural networks, then this book is not for you.

- If you're looking for a book that develops in detail a new theory about the nature of meaning, then this book is also not for you.
- If you're looking for a complete theory of interpretable AI then, unfortunately, you've also bought (or borrowed or downloaded) the wrong book.

Our goals are modest. We hope the book will help frame some important issues that we find surprisingly little literature on. AI raises very interesting philosophical questions about interpretability. This book tries to articulate some of those issues and then illustrate how current philosophical theories can be used to respond to them. In so doing, it presupposes some knowledge of philosophy, but not very much. Our hope is that it can be used even by upper-level undergraduate students and graduate students not expert in either AI or philosophy of language. We hope it will inspire others to explore these issues further. Finally, we hope it opens up a door between researchers in AI and the philosophy of language/metaphysics of content.

# ALFRED (THE DISMISSIVE SCEPTIC)

## *Philosophers, Go Away!*

In the previous chapter, we outlined a range of interesting philosophical challenges that arise in connection with understanding, explaining, and using AI systems. We tried to make the case that philosophical insight into the nature of content (and the difference between a system having content and simply being evidence of some sort) is centrally important both for understanding AI and for deciding how we should integrate it into our lives.

When we first started work on this project, we got in touch with people in the AI community. We thought that our work on these issues should be informed by people working in the field—those who are actually developing ML systems. When we approached people, they were friendly enough. We had many helpful conversations that have improved this book. However, it soon became clear to us that the people working at the cutting edge of AI (and in particular those working for the leading corporations) considered us, at best, lunchtime entertainment—a bit like reading an interesting novel. There was a sense that no one really took these philosophical issues *seriously*. Here is a caricatured summary of the attitude we encountered:

> Look, these big-picture concerns just aren't where the action is. I don't really care whether this program I'm working on is 'really a malignant mole detector' in any deep or interesting sense. What I care about is that I'm able to build a program that plays a certain practical role. Right now I can build something that's pretty decent at the role. Of course there's a long way to go. That's why I spend my time thinking about how adding back propagation, or long short term memory, or exploding gradient dampening layers, or improved stochastic gradient descent algorithms, will lower certain kinds of error rates. If you have something actually helpful to say about a piece of mathematics that will let me lower error rates, or some mathematical observations about specific kinds of fragility or instability in the algorithms we're currently using, I'm happy to listen. But if not, I'm making things that are gradually working better and better, so go away.

We take this dismissive reaction very seriously and much of this book is an attempt to reply to it. We are not going to dismiss the dismissal. At the end of the book, we have not refuted it. There's something to it, but, we argue, it's an incomplete picture and we outline various ways in which it is unsatisfying.

It's worth noting that this pragmatic-sceptic's dismissal of philosophy has analogues in almost all practical and theoretical domains. Practising mathematicians don't worry much about the foundations of their disciplines (they don't care much about what numbers are, for example). Politicians don't care much about theories of justice (they don't spend much of their time reading Rawls, Nozick, or Cohen). Those making medical decisions with massive moral implications don't spend much time talking to moral philosophers. And so it goes. There's a very general question about what kind of impact philosophical reflection can have. One way to read this book is as a case study of how philosophers should reply to that kind of anti-philosophical scepticism.

We should, however, note that not all those working in AI share Alfred's dismissive attitude towards increased reflection on the foundations of ML systems. In 2017, Google's Ali Rahimi gave a talk where he compared the current state of ML systems to a form of *alchemy*: programmers create systems that work, but they have no real, deep, understanding of *why* they work. They lack a foundational framework. Rahimi said:

> There's a self-congratulatory feeling in the air. We say things like 'machine learning is the new electricity.' I'd like to offer an alternative metaphor: machine learning has become alchemy.[1]

It's become alchemy because the ML systems work, but no one really understands *why* they work the way they do. Rahimi is not entirely dismissive of making things that work without an understanding of why it works: 'Alchemists invented metallurgy, ways to make medication, dying techniques for textiles, and our modern glass-making processes.' Sometimes, however, alchemy went wrong:

> …alchemists also believed they could transmute base metals into gold and that leeches were a fine way to cure diseases. To reach the sea change in our understanding of the universe that the physics and chemistry of the 1700s ushered in, most of the theories alchemists developed had to be abandoned.

---

[1] https://www.youtube.com/watch?v=x7psGHgatGM.

More generally, Rahimi worries that when we have ML systems that contribute to decision making that's crucial both to individuals and to societies as a whole, a foundational understanding would be preferable.[2]

> If you're building photo sharing services, alchemy is fine. But we're now building systems that govern health care and our participation in civil debate. I would like to live in a world whose systems are built on rigorous, reliable, verifiable knowledge, and not on alchemy.

Many in the AI community dismissed Rahimi's pleading for a deeper understanding. Facebook's Yann LeCun replied to Rahimi saying that the comparison to alchemy was not just insulting, but wrong:

> Ali complained about the lack of (theoretical) understanding of many methods that are currently used in ML, particularly in deep learning. Understanding (theoretical or otherwise) is a good thing…But another important goal is inventing new methods, new techniques, and yes, new tricks. In the history of science and technology, the engineering artifacts have almost always preceded the theoretical understanding: the lens and the telescope preceded optics theory, the steam engine preceded thermodynamics, the airplane preceded flight aerodynamics, radio and data communication preceded information theory, the computer preceded computer science.[3]

We will argue that Rahimi is right: the current state of ML systems really is a form of alchemy—and not just for the reasons Rahimi mentions. The one important reason is that the field lacks an understanding of how to describe the content of what it has created (or how to describe what it has created as deprived of content). We are presented with AI as if it is something that can talk to us, tell us things, make suggestions, etc. However, the people making AI have no theory that justifies that contentful presentation of their product. They have given us no rational argument for that contentful presentation. They've just written some algorithms and they have no deeper understanding of what those pieces of mathematics really amount to or how they are properly translated into human language or affect human thoughts. If the view is that these programs have no content at all, then that too is a substantive claim that needs justification: What is content such that these systems don't have it?

---

[2] Note: Rahimi's worry is not specifically about interpretability, but the same point applies.

[3] https://www2.isye.gatech.edu/~tzhao80/Yann_Response.pdf.

So: welcome to the world of philosophy. It's a world where there's very little certainty. There are many alternative models, the models disagree, and there's no clear procedure for choosing between them. This is the kind of uncertainty that producers and consumers of AI will have to learn to live with. It's only after a refreshing bath in philosophical uncertainty that they will start to come to grips with what they have made.

## A Dialogue with Alfred (the Dismissive Sceptic)

**Alfred**: I appreciate the interest you philosophers have in these issues. It's important that a broad range of disciplines reflect on the nature of AI. However, my job is to make exactly the kinds of AI systems that you talk about in the introduction of this book and I don't get it. I just don't see that there's anything you philosophers can tell me about interpretation that will help me do my job. We've made all these amazing advances, and we did it without you. I'm not doubting that there's some interesting meta-reflections around these issues, but that's just lunch entertainment for us. It makes no real difference to what we do, day to day. Issues about the nature of interpretation and the nature of content don't seem pressing to me in my professional life.

So, as a conversation starter, let me try this: philosophical theories of meaning and language make no difference to what we do. For professional purposes, we can ignore them.

**Philosopher**: I don't see how you can avoid those issues. What do you think is going on with SmartCredit, then? We give the software access to Lucie's social media accounts, and it spits out the number 550. But so far, that's just pixels on a screen. The output of the program is useless until we know that 550 *means* a high risk of default. We need to know how to look at a program and figure out what its outputs mean. That's absolutely central to our ability to make any use of these programs. We can't just ignore that issue, can we?

**Alfred**: Of course we say things like, 'That output of 550 means that Lucie is a high risk of default.' But that's just loose talk—we don't need to take it seriously. All that's really going on is this. SmartCredit is a very sophisticated tool. It takes in thousands of data points and sorts and weighs them using complicated and highly trained mathematical algorithms. In the end SmartCredit spits out some number or other. That number doesn't in itself *mean* anything. It's just a number—just the end product of millions of calculations. Of course the bank should then take that number into account when deciding whether to extend a loan to Lucie. But not because the number *means* that Lucie is a default

risk—rather, because the number is the output of a highly reliable piece of software.

**Philosopher**: Wait, I'm not sure I understand what you're proposing. Just recently I went to the doctor and he used a machine learning program called SkinVision to evaluate a mole on my back.[4] According to him, SkinVision said that the mole was likely to be malignant, so he scheduled surgery and removed it. Are you telling me that the doctor was wrong and that SkinVision didn't say anything about my mole? I guess then I had surgery for no reason. Or what about the case of Eric Loomis? Loomis was found guilty of participating in a drive-by shooting, and was sentenced to six years in prison in part because, according to the judge, the machine learning program COMPAS said that Loomis was a high risk to reoffend.[5] Are you telling me that the judge was wrong and that COMPAS didn't say anything about Loomis's recidivist risk? If that's right, surely it was a huge injustice to give Loomis more prison time. It looks like we're treating these programs *as if* they are saying things all over the place, and making many important and high-stakes decisions based on what we think they are saying. If that's all wrong, and the programs aren't really saying anything, don't we need to do some serious rethinking of all of this technology?

**Alfred**: I think you're making a mountain out of a molehill here. Again, it's just loose talk to say that *COMPAS says that Loomis is high risk* or to say that *SkinVision says that your mole is probably malignant*. But that doesn't mean we're taking important actions for no reason. *SkinVision* didn't say that your mole was probably

---

malignant, but your doctor did say that. He said it in a sloppy way—he used the words 'SkinVision says that your mole is probably malignant'—but we don't need to take his exact phrasing seriously. It's clearly just his way of telling you (himself) about your mole. And there's no worry about having a mole removed because your doctor says that it's probably malignant, is there? The same with COMPAS. COMPAS didn't say that Loomis was high risk—the judge did. Again, he said it in a sloppy way, but we all know what's going on. And there's nothing wrong with giving someone a severe sentence because a judge says that he's a high recidivism risk, is there? That kind of thing happens all the time.

**Philosopher**: That's helpful. So the idea is that all the meaning and content is in the people saying things in response to the programs, not in the programs themselves. That's why we don't need a theory of content for the programs. (Hopefully we can get a good theory of content for people—but in any case that's not a special problem for thinking about AI systems.) But I'm still worried about how this idea is going to be worked out. My doctor gives SkinVision a digital photograph of my mole, and it produces a printout that says 'Malignancy chance = 73%'. Then my doctor says that my mole is probably malignant. On your view, SkinVision didn't say anything, and its printout didn't have any content—all the saying and all the content is coming from the doctor. But it sure seems like quite a coincidence that there's such a nice match between what my doctor *really meant* and what the words printed by SkinVision *seemed to me* but (on your view) didn't really mean.

**Alfred**: Of course it's not a coincidence at all. The designers of SkinVision included a helpful user interface so that doctors would know what to say when they got the results of a SkinVision analysis. There's nothing essential about that—SkinVision could have been designed so that it just outputs a graph of a function. But then doctors would have needed more training in how to use the program. It makes sense just to have the programmers add on some informative labelling of the outputs on the front end and save the doctors all that work.

**Philosopher**: 'Informative labelling'—I like that. You can't have informative labelling without information. Doesn't that then require that the outputs of SkinVision do mean something, do carry the information that (for example) the mole is probably malignant?

**Alfred**: Good point. OK, what I should have said was not that it's the *doctor* who's the one who's really saying something—rather, it's the *programmer* who's really saying something. When SkinVision prints out 'Malignancy chance = 73%', that's the programmer speaking. She's the one who is the source of the meaning of those words. They mean what they do because of her programming actions, not because of anything about the SkinVision program itself. SkinVision is then just a kind of indirect way for the programmer to say things. That's a bit weird, I admit, but there are lots of other forms of indirect announcement like that. When the programmer writes some code which, when run, prints 'Hello World', it's the programmer, not the program, who greets the world. SkinVision and other AI systems are just more complicated versions of the same thing. The doctor then *also* says that your mole is probably malignant, but that's just the doctor passing on what the programmer indirectly said to him.

**Philosopher**: That's an interesting idea. But I'm worried that it has a strange consequence. Suppose that the programmer of SkinVision had been in a perverse mood when programming the final user interface, and had set things up so that the mathematical output that in fact leads to SkinVision printing 'Malignancy chance = 73%' instead caused SkinVision to print 'Subject is guilty of second degree murder'. Would that then mean that SkinVision, rather than a piece of medical software, was instead a bit of legal software, making announcements about guilt or innocence rather than malignant or benign statuses?

**Alfred**: What? Of course not. Why would you even think that? SkinVision's whole training history shaped that neural network into a medical detector, not a legal detector. How would a perverse programmer implementing perverse output messages change that?

**Philosopher**: Well, doesn't it follow from what you said? If SkinVision itself isn't really saying anything, and it's just a tool for letting the programmer speak, then if the programmer chooses to have it produce the words 'Suspect is guilty of second degree murder', what's said (by the programmer, through the program) is that the suspect is guilty of second degree murder. And if the information conveyed is legal, rather than medical, then it looks like a piece of legal software.

**Alfred**: Not a very good piece of legal software! The guilt and innocence announcements it produces aren't going to have anything to do with whether the person is really guilty. You can't tell guilt or innocence from a photograph of a mole. And even if you could, SkinVision hasn't been trained to do so.

**Philosopher**: Agreed, it would be a terrible piece of legal software. But my point is just that that's what it would be, since its outputs mean what the programmer wants them to mean. I can see that in this case there's some plausibility to the claim that when the perversely programmed SkinVision prints 'Subject is guilty of second-degree murder', what's said is that the subject is guilty of second-degree murder. (Whether it's SkinVision itself or the programmer who's saying this is less clear to me.) But I'm worried that that's a special feature of this example. In this particular case, the programmer has decided to put the program output in the form of words in a pre-existing language. It's thus very tempting to take that output to mean whatever those words mean in the language. In the same way, if a monkey banging on a keyboard happens to type out 'To be or not to be, that is the question', we might feel some inclination to say that the monkey has said something. But probably that feeling should be resisted, and we should just say that the *sentence* means something, and that the monkey has accidentally and meaninglessly produced it.

Consider another case. StopSignDetector is another machine learning neural net intended to be used in self-driving autonomous vehicles. The plan for StopSignDetector was, not surprisingly, to have it be a stop sign detector, processing digital images from a car camera to see if there is a stop sign ahead. But StopSignDetector doesn't print out 'There is a stop sign', or anything like that. There's just a little red light attached to the computer that blinks when the program reaches the right output state. As I understand your view, the blinking red light doesn't mean anything in itself, but is just a device for the programmer saying that there is a stop sign. That's because, I guess, the programmer intends the blinking red light to announce the presence of a stop sign. But now add in the perverse programmer. What if the programmer decides instead that the blinking red light should announce the presence of a giraffe—but doesn't change anything in the code of StopSignDetector. Does that mean that we end up with a very bad giraffe detector?

**Alfred**: I think all of this is getting much more complicated than it really needs to be. We speak sloppily as if these programs are saying things, producing outputs that somehow represent specific facts about the world. That's all just sloppy speech. In many cases, that sloppiness can be fixed up by taking us *really* to be talking about what the end user (like the doctor or the judge) is saying or what the original programmer is saying. But sure, I agree that in weird cases when end users or programmers have weird secret plans, that's not a good way to fix up our sloppy talk. But it's not that hard to find a different way, is it?

Think about your standard pocket calculator. You push the buttons '58 + 67' on the keyboard, and on the display it shows '125'. Does that mean that the calculator is *saying* that 58 plus 67 is 125? Surely not—there's no need for that kind of content talk. Of course, someone using the calculator might then say '58 + 67 = 125', and thereby mean (as people do) that 58 plus 67 is 125. And it's presumably not an *accident* that the calculator display looks the way it does—the original programmer of the calculator software chose that display format because of their plan that the calculator be a tool to announce arithmetic facts. But even if we discovered that the programmer had strange secret plans and the calculator user had strange secret interpretive ideas, it wouldn't matter. That's because in the end the calculator is just a tool for getting at mathematical results. So long as the calculator is working correctly, who really cares what anyone's communicative plans are, or what the calculator or anyone else is 'really saying'.

**Philosopher**: But I'm not sure a calculator is the right comparison for you. The programming of a calculator is a straightforward example of symbolic representational programming. If we look into the coding details of the calculator, we will indeed be able to find the parts of the program that represent numerical values, and that represent the applications of various mathematical operations to those numerical values. Here, it looks entirely natural to me to say that the calculator display really does mean that 58 plus 67 is 125. None of the special features of (for example) SmartCredit that made its contents so obscure seems to be present in this case.

**Alfred**: OK, fair enough. But I bet I could program up a machine learning pocket calculator if I really set my mind to it. I bet you haven't actually checked out the coding of your TI-Nspire—would you really change anything in how you used the calculator if you discovered that it had a neural network implementation?

**Philosopher**: Probably not. But that's because I would think that, whether neural network or not, the calculator's program was *about* mathematical operations. Remember, I'm not a sceptic about the role of content in these cases, you are. I'm happy to say that we don't need to worry about obscure communicative plans on the part of the programmer or the user, because I'm happy to say that the program itself means something. (Of course, I think it's a very hard question *why* it means something, and I think in some cases we might have a lot of trouble figuring out *what* it means.) So what's your view on this? Don't you need a view on what it means to say that the calculator is a 'tool for getting at mathematical results'? That looks an awful lot like a disguised claim about the contents of the calculator claims.

**Alfred**: That's got to be too fast. A hammer is a tool for pounding in nails, right? That's not a claim about the meaning or content of a hammer. That's just an observation about what hammers are useful for.

**Philosopher**: Agreed. But I think this overlooks an important distinction. A hammer isn't an informational tool. When we use a hammer, we're not trying to learn anything—we're just trying to get something done (get some nails in some wood). It's not too surprising if we don't need any notion of content to explain that kind of tool. But a language is also a tool, isn't it? And to say *that* kind of tool, we need to talk about contents. That's because language is an informational communicative tool, a tool that we're using to learn things. So we need to say what sentences mean to see what we can learn from them. And SkinVision and COMPAS look like tools of the same sort. We're not trying to *do* something with those tools—all of the *doing* is by the doctor or the court. We're just trying to get some information out of the tools. And if we're going to get information out, we need a contentful interaction with the program.

**Alfred**: Good, that helps me see what I want to say. In the end, the tools I want to make are more like hammers than like languages. Consider an example. I want to build a self-driving car. I'm not trying to make a car that I'll learn something from—I just want a car that will do something for me. I want a car that I can get into and that will then take me to the right place. That's a big project, so I'm not trying to do it all at once. Along the way, I produce a machine learning image recognition program that will beep when there's a pedestrian in the road. For now, that can be a helpful signal to the driver. But eventually, I'll have that bit of programming integrated into a larger autonomous vehicle program. Once that's all done, all I care about is that the car won't in fact hit pedestrians. Whether the beeps from that one part of the program 'mean that there's a pedestrian in the road' makes no difference to me. Why would I care? I'm not trying to give anyone any information with that beeping; I'm just trying to make sure that the car doesn't crash.

**Philosopher**: I see the idea, but how does that help with other cases? Maybe we don't need to assign contents to the full self-driving car, but before the pedestrian detector is integrated into the full car, while it's being used to warn human drivers, don't we need its beeps to *mean* that there's a pedestrian in the road?

**Alfred**: I don't see why. I'm happy to think of the driver in the same way that I think of the self-driving car. I'm not interested in getting any particular *contents* to the driver. What I care about is that the driver swerves when the program beeps. So long as that happens, and the pedestrian isn't hit, I'm happy.

**Philosopher**: I see. So you're just thinking of the programs as little causal prods that push people into the right kind of activity. SkinVision just needs to cause doctors to perform surgeries; never mind what the doctors believe. COMPAS just needs to cause judges to issue severe sentences; never mind what judges might learn from COMPAS.

**Alfred**: Right. Sure, probably the best way to get doctors to perform surgeries under the right conditions is to get them to believe that people need surgeries under those conditions. But that's just an accidental feature of doctors making them different from nails. The thing that really matters is just that our program causally prompts the right things to happen.

**Philosopher**: I'm not sure this 'it's all just causal prods' idea is going to be as easy to work out as you seem to think. You said you just wanted 'a car that I can get into and that will then take me to the right place'. But where did this notion of 'right place' come from? That requires that the car takes you where you want to go, and that then requires that you are able to *tell* the car where to go. But doesn't that still require a contentful interaction with the program? Maybe it's on the input side rather than the output side, but the issues seem to me to be the same—I need to be able to do something to the program that I can count on putting the program into the right state. I need to be confident that when I tell the self-driving car to take me to the airport, its subsequent driving will be guided by the content of what I told it.

**Alfred**: I'm tempted to say that the problem you're pointing out is just another artefact of our being only part-way through the overall project. I already agreed that *for now* I want the pedestrian detector's beeps to be understood by human drivers as signalling that there is a pedestrian in the road. We're talking about understanding and meaning here because the programming project isn't finished yet, so we can't just let the car do its own self-driving business. But the same is true for the need to give the car directions. Down the road, the goal should be a car that you don't need to give directions to. The car will figure out how to deal with pedestrians in the road; it will also figure out how to deal with a passenger in the car. Maybe it will access your calendar and determine where you ought to be and automatically take you there.

**Philosopher**: Wait, 'figure out'? 'Determine'? 'Access your calendar'? That all looks like content-based talk.

**Alfred**: Sure, but it's all dispensable in the same way. When I say that the car will figure out how to deal with pedestrians in the road, I just mean it won't hit pedestrians in the road. When I say the car will figure out how to deal with a passenger in the car, I just mean that it will take that passenger to a location where the passenger ought to be. And so on.

**Philosopher**: I'm not sure I like the vision of the AI future you're sketching here. These days when I get in the car and drive somewhere, I have plans and reasons for what I'm doing and I perform a bunch of deliberate intentional actions in pursuit of my goals. Your self-driving car takes that all away from me. I don't need any plans, or any reasons for going anywhere. I just get in the car, and the car takes me somewhere that will work out well for me. It feels like a Wall-E future, with all of us passive passengers on the Axiom. It's important to us that we have reasoned engagement with the world—aren't you proposing to shrink that reasoned engagement down to nothing, by embedding us in a network of devices that just causally push us around to where we ought to be?

**Alfred**: Well, as long as you're really getting where you ought to be, is it really that bad? We're surrounded by lots of systems and devices that take care of our needs without our reasoned engagement. When you're exposed to germs, your immune system just takes care of it for you—it causally pushes bits of your body into the right places without any intervention by you. Things wouldn't be any better if you had to reason your way through a viral infection, would they?

**Philosopher**: Fair enough, although just because something is good in some places doesn't mean it's good everywhere. But surely there's also a real issue about whether we can count on the self-driving car taking us where we *ought* to be. What's our source of confidence in that 'ought'? Either we're just building into the program what the right final goals are (get us where our calendar says we ought to be), in which case it looks like we still need content tracking with the program. Or we've got the kind of advanced AI that has the ability to reshape the categories it's been trained to track, in which case, if there's no notion of content of the program's reshaped categories, I'm not sure why we should be confident that what it's doing is in any sense getting us where we *ought* to be.

**Alfred**: Look, all of this is getting extremely speculative. Forget about this utopian/dystopian picture in which our AI systems just shepherd us through the world. Remember, I've already observed that you can think of human users now as being like the eventual self-driving car. I don't care whether the car *knows* that there's a pedestrian in the road and *takes that into account*. All I care about is that when the pedestrian detector beeps, the car changes course. And similarly for the human user. I don't care whether the human user *knows* that there's a pedestrian in the road and *takes that into account*. All I care about is that when the pedestrian detector beeps, the driver changes course. Who cares what the underlying mechanism is by which that happens?

**Philosopher**: There's a sense in which I agree with all of that. Forget about programs entirely, and just think about people. There's some sense in which all of the content talk we go in for may be optional. Maybe we can stop thinking about other people as creatures having beliefs and desires and plans with contents and making claims with contents, and just think about them as lumbering obstacles to be manipulated and manoeuvred around. But surely *something* is gained by instead thinking about people as bearers of content. If we've at least reached the point, then, of saying that content talk for AI systems is exactly as dispensable as content talk for people, I think we've got enough to motivate some careful thinking about how to make that content talk work out in the AI case.

**Alfred**: Fair enough. Let's at least see what you've got.

# A PROPOSAL FOR HOW TO ATTRIBUTE CONTENT TO AI

# TERMINOLOGY

*Aboutness, Representation, and Metasemantics*

We didn't refute Alfred's content-scepticism, but at least we got him to agree to explore strategies for attributing content to AI. That's the goal of the rest of this book. As we pointed out to Alfred, the most salient prima facie argument for doing that is that AI is presented to us, by its producers, as having content. AI systems are presented as *saying* things, as *making suggestions*, and sometimes *making decisions*. We, the human users, typically treat them as conversation partners and as sources of information (contentful information, that is).

There should be no disagreement about the fact AI systems are often (and arguably typically) presented to end-users in this way. There are indefinitely many illustrations of this from academic work to advertisements. Here's a tiny collection of the kinds of claims we have in mind (all emphasis ours). From academic papers:

> SmartBot can fall into false or impossible *beliefs*. For example, SmartBot can *believe* one of its cards has no valid value as all possible cards are inconsistent with the observed play according to SmartBot's convention. (Bard et al 2019)[1]

[1] https://arxiv.org/pdf/1902.00506v1.pdf.

Robots must *know how* to be gentle when they need to interact with fragile objects, or when the robot itself is prone to wear and tear.

(Huang et al. 2019)

For a given set of desired performance measures, i.e. cycle time, work-in progress, and utilisation of three different testers, the neural network *suggests* a suitable design of scheduling rules, and the number of each type of tester needed to achieve management's goal.   (Alam et al. 2004)

It is W [weights linking nodes] that constitutes what the network *knows* and determines how it will respond to any arbitrary input from the environment.   (Tam 1991)

From more technical computer science blogs:

Neural networks are a set of algorithms, modeled loosely after the human brain, that are designed to *recognize* patterns. They *interpret* sensory data through a kind of machine perception, labeling or clustering raw input. The patterns they recognize are numerical, contained in vectors, into which all real-world data, be it images, sound, text or time series, must be translated…

A neural network is a corrective feedback loop, rewarding weights that support its correct *guesses,* and punishing weights that lead it to err.   (Nicholson, skymind.ai)[2]

From a more or less general interest computer science blog:

DeepXmas: AI *knows* if you are naughty or nice…

This AI home security system can use deep-learning and *figure out* when kids are making messes, or doing things that need action. This type of technology could save a life in the future (e.g. kid choking on a blind cord). ("We don't want the AI to just become a person or kid detector, we want it to understand *naughtiness.*")[3]

---

[2]  https://skymind.ai/wiki/neural-network.
[3]  https://towardsdatascience.com/deepxmas-ai-knows-if-you-are-naughty-or-nice-2bd00b2ad3d2?gi=f35896d7ff04.

# Loose Talk, Hyperbole, or 'Derived Intentionality'?

Recall from the previous chapter that Alfred dismissed all of this as *loose talk* or *hyperbole*. He thinks it is useful, but fundamentally misleading jargon. Maybe it is not just misleading, but also false. According to the sceptic, AI can't, literally speaking, perform speech acts, nor can AI think, believe, or have any of the mental states that we humans have. The sceptic says we find it natural to say things like:

The calculator *says* that 87x9= 783,

but calculators don't really say anything. Our theory of what it is to perform the speech act of saying need not account for calculator-speech. That ordinary speech is filled with false anthropocentric descriptions of calculators shouldn't mislead philosophers. The same goes for talk about the kinds of advanced AI we focus on in this book.

It is very important to emphasize that we don't take ourselves to have refuted that kind of view. Our goal is a more modest one: as a counterbalance to that dismissive view, we will consider some first steps towards making (some) content attributions true. The best way to work out an alternative to the no-content view is actually to try to work out some of the details of an alternative.

Here is a reason for being a bit interested in our effort. Start by asking why we take content talk more seriously for people than we do for calculators. It's because people act in lots of complicated ways that the content attributions help make sense of, while the calculator doesn't really have complex actions (a purely physical account of what's going on with circuits in the calculator lets us understand what we want to understand). But the AI systems that we are considering also act in very complicated ways that aren't illuminated by looking at what's going on with the microstructure. It is particularly interesting that their complicated forms of behaviour aren't the same as the human complicated forms of behaviour. This is what we need to engage in: what we below will call *anthropocentric abstraction*.

None of this amounts to a conclusive proof that certain forms of AI can perform certain forms of speech acts. However, note that if, after reading this book, you end up finding our efforts unconvincing, you'll do so because you have philosophical arguments against what we say. You have in effect used theories about the metaphysics and methodology of content attribution to help you understand AI and our interactions with AI. That's support for one of our central messages: there should be more interaction between theories about the metaphysics of content and theories of AI.

## Aboutness and Representation

Compare two things. On the one hand, a sock—say, the left sock worn by Alan Turing when he started writing his famous paper on computability—and on the other, this sentence:

(1)   The Eiffel Tower is in Paris

The sock is not about anything. It just exists. It can be worn, washed, and mended, but it doesn't represent anything. The sentence 'The Eiffel Tower is in Paris', on the other hand, *is* about something—it exhibits what philosophers imaginatively call 'aboutness'. English speakers have an easy time identifying what it is about: most obviously, it is about Paris and the Eiffel Tower. It's about those two objects because 'Paris' is the name of Paris and 'the Eiffel Tower' is the name of the Eiffel Tower. However, it is not just about those two objects. It is also about the latter being located in the former. On one view, for example, there's something called a fact or a situation, *that the Eiffel Tower is in Paris*, that the sentence represents.[4]

A striking fact about (1) is that by virtue of its representational properties, it is the kind of thing that can be true or false.[5] If the Eiffel Tower is in Paris, then (1) is true. If the Eiffel Tower is not in Paris, then (1) is false. As it happens, it has the property of being true. Another way to put this: (1) represents the world as being a certain way. If the world is that way, then (1) is true. If the world is not that way, then (1) is false. The world is that way, so it is true. The sock, on the other hand, cannot be true or false. It doesn't represent the world as being any way at all.

The phenomenon of aboutness is so familiar to us (and so central to our lives) that it is easy to forget or overlook how amazing it is. We just used the word 'Paris', sitting in Oslo, and somehow that word manages to 'reach' all the way to a physical structure 1,555 km away (and does so without a passport or a plane ticket). Somehow the word 'Paris' connects with Paris. Not only can aboutness cross space in a seemingly mysterious way, it can also cross time. The expression 'Emperor Kanmu' denotes a Japanese emperor who lived more than 1,000 years ago. Just having read the previous sentence, you, our reader, can now use the expression 'Emperor Kanmu' to *talk about Kanmu.*

---

[4] For introductory work on the metaphysics of facts, see Armstrong (1997) and Mulligan (2007).

[5] We sidestep issues as to what the most fundamental 'truth-bearers' are because they aren't immediately relevant.

A sentence like (1) is an artefact. It consists of objects that we have constructed, i.e. words. But it is not only artefacts that have aboutness. We humans can believe, think, hope, fear, expect, conjecture, etc. In so doing, our minds are directed at features of the world in much the same way sentences of a language are. Just as you can say, in English, that the Eiffel Tower is in Paris, so you think, or believe, or hope, or fear that the Eiffel Tower is in Paris. In the nineteenth century, the philosopher Franz Brentano (Brentano 1874) introduced (although cognates, meaning similar things, had already existed in, for example, Latin) the term 'intentionality' to denote this ability of the human mind to represent. In the current literature, the terms 'intentionality', 'representation', and 'aboutness' are often used interchangeably (though in some theoretical contexts they are distinguished). In what follows we'll for the most part use 'representation', sometimes 'aboutness', and leave 'intentionality' behind.

## AI, Metasemantics, and the Philosophy of Mind

There is a vast literature spanning many subdisciplines of philosophy that attempts to give an account of what representation amounts to and how it comes about. For more than 100 years, philosophers have been concerned with questions such as:

- By virtue of what can a *sentence of English*, say (1) above, be about the Eiffel Tower?
- By virtue of what is *the thought that the Eiffel Tower is in Paris* about Paris?
- What is the connection between the answer to those two questions?

When applied to language, these kinds of questions are often described as parts of *metasemantics*.[6] When the focus is on the intentionality of the human mind, the relevant literature is often classified as philosophy of mind.

To put this book into perspective, it's worth noting that most of the contributions to the philosophy of AI have drawn on work done in the philosophy of mind. It has not been based on work done in the philosophy of language (or the intersection of philosophy of language and philosophy of mind), and not paid any heed to the externalist tradition in the philosophy of language that is the theoretical foundation both of this work and of much of the most important work in twentieth-century philosophy of language and mind. It is hard to prove a negative, but the reader could look at the bibliographies of recent overview works—of Bringsjord and Govindarajulu (2018) on the philosophy of AI, or—perhaps more pertinently—Bruckner (2019) for the philosophy of deep learning. We see literally no works of philosophy of language there. The same thing applies to an extensive review of social sciences (including philosophy) on explainability and AI (Miller 2018). The closest engagement we've found with philosophy of language is in the monograph Floridi (2011), but in that work there is no Burge, no Kripke, no Putnam, no externalism. In the same vein, the otherwise excellent book *How to Build a Brain* (Eliasmith 2013) is written as if the externalist tradition doesn't exist.[7] Other recent work approaches the topic from a functionalist perspective (López-Rubio 2018); from a Kantian perspective (Schubbach forthcoming); from a teleosemanticist perspective (Shea 2018) or using sui generis theoretical tools (Floridi, the book just mentioned); through the lens of more venerable philosophical conceptions of abstraction as found among the British empiricists and their followers (Buckner 2018); or from the perspective of modern compositional semantic theories (Nedft 2020). In all this we find no mention of the externalist tradition, a strange gap in the literature we're aiming here to fill.

---

[6] See, for example, Brentano (1911) and Crane (1998) for people who take aboutness to be fundamental to theory of mind, the language-of-thought theorists like Fodor (1975), and the teleosemanticists like Dretske (1980) and Millikan (1984). Classic works on the representational properties of language include Russell (1905), Strawson (1950), Kripke (1980) (discussed at length below), Donnellan (1966), and Evans (1982). More recent work that shows these issues remain live concerns includes Recanati (2012) and Hawthorne and Manley (2012), and a textbook introduction that brings one right up to date on active issues in the field is Cappelen and Dever (2018).

[7] There is of course significant work on the extended mind hypothesis (Clark and Chalmers 1998) and while this is a form of externalism, it is not one based in the Kripke, Burge, Putnam tradition. There is also work on what is called 'embodied embedded cognition', and while this could also be called a form of externalism, it is entirely different from the tradition we are relying on here.

# 4

# OUR THEORY

## *De-Anthropocentrized Externalism*

Our goal in this book is to point in what we think is the right direction for explaining the contents of AI systems, and to do some initial exploration of the territory in that direction. In this chapter, we'll set out two central claims (and a third peripheral claim) that structure our positive proposals. The first claim is directed primarily at work being done in the artificial intelligence literature. That claim contains a bit of bad news: much of the work being done on interpretability of artificial intelligence, we think, centres around an incorrect picture of how content is determined. But it also contains a bit of good news: work on the determination of content in philosophy provides a better *externalist* picture of how content is determined and sophisticated tools for developing that picture.

The second claim is directed primarily at philosophers. The second claim also starts with a bit of bad news: philosophers shouldn't get too triumphant about the special suitability of externalist theories of content determination for the AI content project. That's because consideration of some of the specific details about AI systems reveals that the externalist accounts that philosophers have developed contain crippling anthropocentric biases that make them unsuitable for use on nonhuman cases like AI systems.

But again there's a bit of good news: consideration of those specific details also helps point the way towards a deeper and more generalized understanding of externalism, one that gives a picture that can apply across human and nonhuman cases.

Putting the two claims together, one of the lessons is that there is room for highly productive interaction between philosophers and artificial intelligence researchers here. Both sides, we think, have been hampered by narrow perspectives. On one side, people have been approaching problems of AI content with an unnecessarily narrow picture of how contents might be determined. On the other side, people have been thinking about content determination using an unnecessarily narrow range of cases of content bearers. Each side has things the other side lacks; bringing everyone together opens up the potential for deeper and more productive work by everyone.

We'll see eventually that when the pieces are brought together, important methodological questions arise about how to carry out a research project using all of those pieces. Our third peripheral claim is then that a characteristically philosophical move will be useful: in order to get the right picture about how to develop a metasemantics of AI content, we need to think first about some *meta*-metasemantic questions.

## First Claim: Content for AI Systems Should Be Explained Externalistically

Machine learning neural net programs are different from other programs in a way that matters for content. Consider a simple bit of Python code:

```
If num > 1:
    for i in range(2,num):
            if (num % i) == 0:
                    print(num," is not a prime number")
                    break
            else:
                    print(num," is a prime number")
    else:
        print(num,"is not a prime number")
```

When the variable 'num' is set to 5009 and the program outputs '5009 is a prime number', there is a clear story about why that output means that 5009 is a prime number. The output means that 5009 is prime because of the programming/computational details about how that output is produced. The program produces the output '5009 is a prime number' because it tries dividing 5009 by all integers between 2 and 5008, and fails to find a nontrivial integer divisor of 5009.[1] Because not having an integer divisor is what it is to be prime, the computational production of the output is representing primeness.

As we've seen, the computational structure of machine learning neural networks makes it incredibly difficult to produce this kind of story grounding content in the programming/computational details. When we look at the vast array of internode connection strengths in the neural network of SmartCredit, or trace the computational path of Lucie's financial details as they percolate through that network, we don't find anything that obviously produces a helpful content-level story. This program-level obscurity then leads people to various reactions:

- Some people get tempted into scepticism about content, thinking that there's no way to tell a content-level story on the basis of such obscure computational mechanisms.
- Some people get tempted into thinking that AI systems must have wildly alien contents, representing (perhaps) massively disjunctive properties that are computationally tracked by the details of the neural network, but which can't be expressed or comprehended by humans.
- Some people think that humanly comprehensible contents can be extracted from the computational details of the neural network, but that sophisticated tools of computational intervention are needed to figure out how specific contents are grounded in specific portions of specific neural networks. (Example: the rapidly expanding body of work on feature visualization tools.)

---

[1] This way of saying what the program is doing takes for granted that, for example, the programming code 'num % i' corresponds to division (mod i). A more complete explanation would explain this representational feature in terms of lower-level architecture.

Our first central lesson for this book is that all of the above is the wrong way to think about the problem of AI content. AI content is not a problem at the level of programming and computational detail. Instead, AI content is a problem at the level of environmental and sociological detail.

One important thought for considering the foundation of content for neural network AI systems is: all of this has happened before. This is not the first time we've encountered the problem of assigning content to systems whose computational details are enormously, and perhaps incomprehensibly, complex. That's because all of us are neural networks of exactly that sort. And we've thought a lot about how to attribute representational content to people.

It's possible to take the computational perspective on content determination for people. To take this perspective involves thinking that what a person means by the sentences they utter, or what the contents of their beliefs are, or what features of the world they are representing in perception, are determined by the computational details of their neuroanatomy. Taken to the extreme, this sort of project leads to identifying representational functions of specific neurons, for example, 'face detection neurons'.[2] Of course, the computational approach doesn't need to be taken to this extreme—we might think that human content is grounded at some 'higher level' of computational organization. Maybe we need multiple neurons working together in the right way before they can represent anything, or maybe we need entire regions of the brain computationally organized in the right way before we get representation. But from this computational perspective, we can extract a research program: consider the computational structure of various parts of human brains at various levels of abstraction, and try to determine which of those computational structures manage to represent and what they represent.

But there's an important alternative perspective on content determination which argues that that entire research program is a mistake. *Externalist* views hold that the problem of content determination for people isn't a computational problem; it's an environmental and sociological problem. (More carefully: externalism holds that content determination isn't *uniquely* a computational program. While a person's internal state may do some of the work in fixing content, their external connections to their environment and society also do some of the work. So in thinking about content determination, we shouldn't look automatically and single-mindedly at computational factors.) There are many versions of externalism available, but what they have in common is a denial that a person's representational capacities are fully grounded by internal features of that person. Instead, externalist views hold that we represent the way we do in part because of features external to us. For example:

- Some of our visual experiences might be representations of faces in part because of our evolutionary history, which is a history of a social species whose survival and reproductive success depended on recognition of social cues, leading to evolutionary selection for facial recognition abilities. On this picture, it's not computational features of our neurons that make them facial detectors, it's historical and teleological features of us that make us facial detectors.
- Some of our linguistic utterances might mean what they mean in part because of the way that we are related to our larger speech community and because of what the words we use mean in that speech community. On this picture, knowing everything about the computational architecture of a particular language user could leave us far short of what's needed to assign content to that user's utterances, since we need at a minimum to know how other people are using those words as well.
- Some of our beliefs have the contents that they do because of the specific environments in which we've acquired

these beliefs. Someone who lives in an environment surrounded by water will acquire beliefs *about water,* while someone who lives in an environment surrounded by some other clear liquid will end up with beliefs about that liquid. (On this picture, looking at computational features of the brains of the believers won't reveal what their beliefs are about—the two believers in the two environments could have their different beliefs despite having the same internal computational organization.)

[2] See e.g. Axelrod et al. (2019).

Those are just some quick snapshots of some externalist approaches—later in this book we'll develop some of the externalist approaches in more detail.

One of our central theses, then, is that we should pursue externalist approaches to content determination, rather than internalist computationally oriented approaches. An analogy to make the point clear:

Suppose we get interested in the question of what makes things *valuable.* So we collect various things of value: dollar bills, krone coins, gold doubloons. We start examining the samples at the microlevel, looking for the features that make them valuable. Looking at the atomic level, we don't find anything clearly value-determining. So we look at a slightly higher level, checking the chemical and molecular structures. Still nothing value-determining appears, so we go up another organization level—we look at threads in the dollar bill, or the ridges on the krone coin. Again, nothing value-determining emerges.

The problem, of course, is that this approach to the determination of value is fundamentally misguided. Value isn't the kind of feature that emerges at some specific physical level of the description of a valuable object. No amount of probing features at the right level is going to produce useful results, because the problem isn't a vertically determined one. Explaining the determination of value needs to be done (at least in part) horizontally, by thinking about how the valuable object is related to us, our practices, and things in the environment. The determination of value is a sociological problem, not a microphysical problem.[3]

Similarly with content, we suggest, Much of the interesting work that's been done in the artificial intelligence literature on interpretability and explainability of AI has presupposed a prob-lematic internalistic and computational perspective, and assumes that the research project needs to be centred around probing kinds of content to be found at various levels of computational organization. Externalist approaches offer much greater promise for explaining AI contents. That shouldn't be surprising. Externalist approaches, since they allow content explanation to be 'horizontal', making it an environmental and sociological ques-tion, rather than 'vertical', let us shortcut full engagement with the computational complexities and obscurities of AI systems. These are the kinds of approaches that have been most successful for 'neural network' creatures like us; it makes sense that they would also be successful for the neural networks we have created. And it's not surprising that content in general would be an externalist notion, grounded in relations of content-bearing creatures and systems to their environment—the nature and role of content, after all, is precisely to relate creatures and systems to their environment, so that they can encode information about the environment and properly interact with the environment.

---

[3] That point is compatible with the thought that the sociological *eventually* gets grounded in the microphysical in a reductionist way, but (a) it doesn't require that further thought, and (b) even if the reductionist agenda eventually works out, it remains true that the right explanatory approach to value goes through the sociological.

## Second Claim: Existing Externalist Accounts of Content Are Anthropocentric

Our first claim is that externalist accounts of content determination provide the right route forward. But it's not enough just to point to existing work in externalist metasemantics. Existing externalist metasemantic stories have been told as stories about people like us, but AI systems, despite some architectural similarities to people, aren't entirely like us. Our first claim started with the thought that all of this has happened before—that the content-determination problems that AI confronts us with are problems we've already encountered in thinking about our own contents. But, as Twain emphasized, history doesn't *repeat*, it *rhymes*. It's the same problem, but in a different key.

Consider a simple case (later discussion in the book will provide more sophisticated discussion of more sophisticated cases; for now, we just want proof of concept). Suppose Jones's visual perceptual experience represents the object in front of him as a snake, and we want a story about *why* snake is part of the content of that visual experience. Why does Jones's visual experience represent a snake, rather than a thin rectangular region with portions of black, red, and yellow? An externalist might at this point appeal to external features of Jones, including his evolutionary history. Jones's visual experience represents a snake because Jones is a member of a species whose visual systems evolved in an environment containing dangerous snakes, so that having snake-recognition capacities was evolutionarily advantageous.

Obviously we aren't going to get explanations quite like that for artificial intelligence systems. Artificial intelligence systems didn't evolve—these programs aren't members of species that reproduce via offspring that mix genetic traits from two parents, they aren't at risk of being killed by predators in their environment before reproducing, they aren't subject to random mutations. More generally, AI systems have very different sorts of environmental and sociological connections than we do—these differences then create problems in taking off-the-shelf externalist tools for content determination, created as theories about human content, and applying them directly to AI systems.

So far we have been discussing externalism at a very high level of abstraction—as the general view that environmental and sociological factors can matter to content determination, and thus that content is not determined solely by the internal computational state of a content-bearer. But to have a substantive theory of AI content determination, we need to descend from that high level of abstraction and say which environmental and sociological factors matter and in what way they matter.

Philosophers have developed impressive models for how to understand the content of human language and human mental states. We have developed theories of what content is, how it can be expressed in language, and how it can be shared in communication. Those theories, however, were developed with humans as their starting point. In other words, we developed those theories by observing a specific animal, with specific biological features and evolutionary history. There are many features of our communicative patterns that are contingent on the kinds of animals we are and the kinds of lives we lead. For example, we have mouths and ears, which make talking and listening natural. We have fingers, which make writing (and sign language) possible. And so on.

However, most people who study meaning and communication will agree that things made very differently from us can express content. If there are aliens, we might be able to communicate with them. They might be able to think things and say things to us, even if both their internal hardware and their external relation to their environment are very, very different from ours. In other words, the ability to communicate in a contentful way is *multiply realizable*: it is an ability that is not restricted to animals and certainly not to animals just like us.

These two facts (that our theories of content have been developed with humans as their starting point but beings other than humans, plausibly, can represent and communicate) are in tension, and failure to attend properly to the second fact has caused our theories to be *biased*. The bias is that most of our theories of representation are too *anthropocentric*. They are parochial because they are based on continent features of our communicative practice. These features are salient to us, but not essential to the nature of content and communication.

Philosophical work in metasemantics, because of its focus on creatures like us, has produced what we will call an **anthropocentric metasemantics**. The existing philosophical accounts of content determination are too parochial by being too focused on contingent features of human communicative/representational practices. What's needed (both for a metasemantic account that's suitable for AI systems and for a general approach to metasemantic questions that's general and robust enough to be philosophically satisfying) is a **de-anthropocentrized** metasemantics. To achieve that de-anthropocentrization, we'll set out the idea of **anthropocentric abstraction**. In anthropocentric abstraction, we take existing externalist accounts of content determination and abstract away from these contingent and parochial features of human communication to reveal a more abstract pattern that is realizable in many kinds of creatures.

The trick with anthropocentric abstraction is that we can't simply abstract away all the details about the specifics of human engagement with environment and society. An abstracted metasemantic theory that said just that content in general (not just for creatures like us) is determined by *some kind of relation to the external environment* would be too vacuous to be useful or interesting. What's needed is to abstract just the right amount: enough to remove any undue anthropocentric bias, but not so much that we remove all content from the externalism.

Finding this abstractive sweet spot will inevitably involve careful consideration of the details of AI systems. We need to consider the points of similarity and difference between AI systems and us, so that we can see how to take externalist frameworks originally developed as tools for understanding our ability to represent the world and abstract them into tools that also explain the ability of AI systems to represent the world. We'll dive into details as we consider some specific externalist frameworks, but we'll start by noting six big-picture points of comparison:

1. **Creation**: Unlike humans, AI systems are intentionally designed and created by people who already have their own representational contents. AI systems thus give rise to special questions about how their contents relate to the contents of their creators.

2. **Limited Range**: Many AI systems have a very limited range of conceptual applications. An image recognition program might only be able deploy the contents 'cat' and 'dog', and only be able to apply those contents to photographic images. Humans, on the other hand, have a very wide range of contents that can be applied across a wide range of domains.

3. **Unclear Boundaries**: Programs, unlike people, easily break down into smaller subprograms, and easily integrate with other programs to create larger computational and functional units. Questions about what exactly has the content are thus trickier for programs than for people.

4. **Output Variability**: Some contentful AI outputs are linguistic, and at least on the face of it, these linguistic outputs have the same content as sentences in a natural language. Other contentful AI outputs are non-linguistic: AI systems can produce numerical outputs (probability distributions), moves on a game board, digital photographical images, and so on.

5. **Dedicated Integration**: AI systems typically have very specific roles that they are intended to play in our lives (assess credit risk, play chess games, etc.), and the contents they bear need to help make sense of them playing those roles. AI systems are largely single-purpose tools; we are largely many-purpose tool users; this difference between us and AI systems can matter to the details of content determination.

6. **Black Box and White Box Implementation**: Like us, AI systems have internal computational architecture that is largely black box, with computational details that are obscure and not revelatory of purpose or representational content. But many AI systems are in fact complicated mixtures of black box and white box components.

A complex neural net might, for example, be combined with a Monte Carlo randomizing tree search algorithm whose computational implementation, purpose, and representational significance are all entirely transparent.

It is in thinking through points of difference and similarity such as these that philosophical work on metasemantics has much to learn from AI research as we look for the right abstractive sweet spot.

## Third Claim: We Need Meta-Metasemantic Guidance

The problems of anthropocentric abstraction are not unique to AI systems. To attribute content to animals, we may need to engage in some anthropocentric abstraction, abstracting away from human-specific details to an approach suitable to the specific ways that other animals are embedded in their environments. The same might be true for content attribution to, for example, pictures, dance, music, and film. In all these cases we might need models that go beyond what we have developed to account for content of human beliefs and languages.

Different cases of anthropocentric abstraction will involve confronting different questions about how to abstract. We need many different metasemantic theories: a human metasemantics, explaining the specific ways in which contents of humans are grounded in specific internal features of humans and specific ways that humans are embedded in their environments; an artificial intelligence metasemantics, explaining the different specific ways in which contents of AI systems are grounded in the different specific internal features of those systems and the different specific ways that they are embedded in their environments, and so on, for other varyingly alien bearers of content.[4]

Each of these domains will require detailed separate investigations: anthropocentric abstraction is not a unified type of theorizing. Here we focus on the problem of abstracting externalism in a way suitable for AI systems. As we'll see in the subsequent chapters, in considering how to abstract, we encounter a number of choice points. As a result, we are exploring a large logical space—a space containing multiple metasemantic theories for different content-bearing creatures and systems, and different options for how to analogize a metasemantic theory for one kind of creature to a metasemantic story for a different kind of creature. Navigating that logical space raises a methodological question: how do we decide what the right way is to abstract? Even once we are completely clear on all of the ways in which AI systems are different from and similar to us, and completely clear on what the right externalist metasemantic framework is for us, how do we decide what abstracted analogue of that framework is best for the AI systems?

Our third central claim in this book is that this methodological problem is best addressed by considering questions

---

[4] We don't mean to commit here to any particular way to carve up the metasemantic landscape. Maybe different kinds of AI require different kinds of metasemantics; maybe humans and some nonhuman animals all get contents determined by the same metasemantics.

of *meta-metasemantics*. Consider an explanatory hierarchy of content-related facts:

1. The **semantic** facts are facts about what contents specific content-bearing items have. It's thus a semantic fact, for example, that the word 'Aristotle' in English refers to Aristotle.

2. The **metasemantic** facts are facts that explain why the semantic facts are what they are. The semantic fact that 'Aristotle' refers to Aristotle is, for example, according to a Kripkean metasemantic approach (which we'll discuss in greater detail in Chapter 6) explained by the fact that the name 'Aristotle' is part of a causal chain of usages going back to Aristotle.

3. The **meta-metasemantic** facts are facts that explain why the metasemantic facts are the way they are. Meta-metasemantic questions are rarely explicitly addressed by philosophers. Why is the semantic fact that 'Aristotle' refers to Aristotle explained by a Kripkean causal chain metasemantics rather than by some other metasemantic account?

Answering meta-metasemantic questions, we will suggest, requires considering the theoretical role of contents. By considering what explanatory work contents and content attributions do for us, we can work out what kinds of fact could best fix semantic features so that contents can play those explanatory roles. A good meta-metasemantic framework can thus offer us the needed methodological guidance. In determining how to abstract an AI-suitable metasemantics from the existing human-targeted externalist metasemantics, we need to think about the role we want contents to play, and then think about the details of AI systems and the role of those AI systems in our lives and our environment. From all of this we can hope to extract a specific picture of what content determination mechanisms for AI systems would be best suited to the roles that the meta-metasemantics identifies.

# A Meta-Metasemantic Suggestion: Interpreter-centric Knowledge-Maximization

The field of meta-metasemantics is less well developed than metasemantics.[5] There isn't even a consensus that metasemantic theorizing should be guided by an explicit meta-metasemantics. One reason for that might be a healthy fear that it is hard to see where this will stop: why not develop the field of meta-meta-metasemantics? After all, if we need the metasemantics to guide our semantics, and we need meta-metasemantics to guide our metasemantics, why and how would this ever stop?

We recognize this as a concern to some, but we have no fear: we endorse this endless hierarchy of theorizing. There is of course a practical limit to what we humans can process and grasp, but that isn't the limit of interesting inquiry. In this book, however, we move at most three meta levels up—we will leave the explorations of higher levels to others (or to ourselves in the future). We do that with a very concrete goal in mind: to guide our theorizing about the metasemantics of AI. Moreover, we will not devote the book to arguing for our meta-metasemantic view. We will instead use a proposal made by Timothy Williamson in the last chapter of *The Philosophy of Philosophy* (Williamson 2007).

---

[5] There is a bit of meta-metasemantic literature on the question of whether meta-metasemantics for externalist theories should be externalist or internalist (see e.g. Cohnitz and Haukioja 2013 for discussion and references).

Before briefly sketching that view, we should emphasize that if you have alternative theories about meta-meta-metasemantics, we encourage exploring the various ways those views will trickle down to metasemantics and that again to particular interpretations of AI output. The overall spirit of this book is to develop a framework for thinking about interpretable AI and there will be many alternative ways to fill in that framework. The use of Williamson's meta-metasemantics is just one of them.

In *The Philosophy of Philosophy*, Williamson can be read as proposing a version of the principle of charity as a meta-metasemantic principle. Roughly, Williamson's view is that the correct metasemantics is one that *maximizes knowledge for the interpretee*. Moreover, Williamson thinks this is the principle that makes externalism correct. His proposal is that a knowledge-maximization principle is the foundation of externalist metasemantics.

The argument goes as follows. Imagine a case of demonstrative misidentification: Alex is a devoted physiognomist who thinks he can tell a person's character from their face. He sees Bea, and on the basis of her appearance forms the belief he would express by saying 'She is F, G, and H'. She is none of these things: physiognomy is nonsense. But, it so happens, there is someone, somewhere (let's say New Zealand), who is F, G, and H, and has no other properties. Call that person 'Ceres'.

We can use this scenario, Williamson thinks, to shed light on our theory of reference, and in particular to draw connections between interpretation and reference. Consider the question of who Alex's utterance (or the related thought) of 'she' refers to.

Does it refer to Bea, the person in front of him, or Ceres, the person who is F, G, and H? If you aim to maximize the true beliefs you impute to someone (à la work like Davidson 1973), taking the reference to be Ceres seems like the way to go: it makes Alex's belief come out true.

But that's obviously wrong: 'a descriptive theory of reference gone mad' (2007: 263), in Williamson's words. The referent, it seems, is Bea. But how do we make that square with interpretation?

Williamson's neat idea is that what we should aim to maximize is not the interpretee's true beliefs, but their knowledge, and that doing so yields an argument for externalism. Thus, to put the matter crudely, imagine the speaker uttering the following four sentences:

- She is F.
- She is G.
- She is H.
- She is in front of me.

If we want to maximize belief, we should say that 'She' refers to Ceres, since that gets us three true beliefs, and one false belief, although it gets us no knowledge. If we want to maximize knowledge, we should say that 'she' refers to Bea, since it gets us one piece of knowledge, namely that Bea is in front of the speaker.

The reason for this is that perception is a suitable 'channel' for knowledge, whereas physiognomy isn't. But perception is of course a paradigm causal channel as well, and so Williamson, generalizing from these sorts of considerations, suggests that knowledge-maximization is a better principle of interpretation than belief maximization because knowledge tracks causal channels in a way that true beliefs don't always. As he intriguingly suggests:

> Such examples [as the one we just considered] are of course just the analogue for demonstrative pronouns of examples Kripke and Putnam used to refute descriptive cluster theories of reference for proper names and natural kind terms. In effect, such theories are special cases of a truth-maximizing principle of charity. One fundamental error in descriptive theories of reference is to try to make true belief do the work of knowledge. (2007: 264)

Our aim here is not to get too deep into theories of interpretation.[6] But we do want to make one change to Williamson's theory: we suggest choosing a metasemantics that maximizes knowledge of the *interpreter*, not of the subject (be it a person or a machine) being interpreted. In other words, we are suggesting our metasemantic principles should be guided by a meta-metasemantic principle that tells us to pick a metasemantics that maximizes what we, the interpreters, end up knowing as a result of the interpretative enterprise.

That raises the question: *why aim for meta-metasemantic principles that tell us to maximize the interpreter's knowledge and not the interpretee?* A couple of points in reply to this: first, note that this is a question in meta-meta-metasemantics. We will not try to provide a general meta-meta-metasemantic theory here. We are not alone in not doing that. In fact, we know of no worked-out three-level metasemantic theory. Since, in a little book like this, arguments have to stop somewhere, we would be fairly comfortable simply using this as a starting point. However, there's a bit more to be said. Recall our earlier point that the meta-metasemantics should be guided by considerations of what explanatory work content and content attributions are doing for us. We think it should be fairly noncontroversial that a central goal of content and content attributions to AIs is to increase our knowledge. That gives us motivation for being self-centred. It is *our* knowledge that matters, not that of the artificial systems.

---

[6] The most sophisticated recent work on the topic is by Robbie Williams (see 2005, 2007) and references therein. A nice overview of the space of options for charity principles is in Felman (1998).

What we just said leaves open the possibility that others might have other interests. The artificial systems, for example, if they have interests, might want to rely on different meta-meta-metasemantic principles when they interpret each other or us. Maybe they want metasemantic principles that maximize knowledge (or power or something else) of the artificial systems. It is also possible that our interests are aligned. That's an open question. More generally, we think the right view is one according to which *interests* of various kinds will play a central role in higher up on the meta-…-metasemantic hierarchy.

These larger questions about how to ground the various levels of metasemantic theorizing are interesting (and worthy of an entire separate monograph, since they are issues largely unexplored), but will not concern us in what follows. Instead, we will use versions of the Williamsonian principle to illustrate how meta-metasemantics can and should play an important role in the z philosophy of AI. More specifically, whenever we engage in de-anthropocentrizing (which is our way of developing a metasemantics for AI), there will be choice points. We will use knowledge-maximization as our guide when making those choices.

# APPLICATION

## *The Predicate 'High Risk'*

In this and the next two chapters, we show how the framework outlined in Chapter 4 can be applied to particular outputs of AI systems. We'll take as our example the output from SmartCredit that resulted in Lucie being turned down for a mortgage. Recall from Chapter 1 that according to SmartCredit, *Lucie is high risk*. We'll split this statement into three parts and present a separate theory for each part:

(i) The statement is about Lucie, i.e. SmartCredit refers to Lucie. We want to figure out how SmartCredit can refer to Lucie.

(ii) The statement is about the property of *being high risk*. We want to figure out how SmartCredit can denote that property.

(iii) The statement predicates the property of being high risk to Lucie. On the assumption that SmartCredit has the capacities outlined in (i) and (ii), we next want to figure out how SmartCredit can attribute the property of being high risk to Lucie.

This chapter is about (ii), i.e. how SmartCredit can pick out the property of *being high risk.* Chapter 6 is about how SmartCredit refers to Lucie. Chapter 7 outlines a proposal for how SmartCredit can predicate *being high risk* to Lucie. Each chapter will make use of different externalist tools: for names we use ideas from Evans; for predicates, we use ideas from Kripke; for predication we use ideas from teleosemantics. The proposals all instantiate the general strategy we presented in Chapter 4.[1]

Our test case is relatively simple. There's an enormous amount more to be done even if we're completely successful with this test case—attributing content to AI outputs that don't have linguistic form, and especially attributing content to non-explicit AI internal states that don't appear as output (understanding SmartCredit not just as denying a loan, but doing so for some reason). However, even the simple test case will be challenging enough for now and so that's where we'll start. Once we succeed with these baby steps, it'll be time to move on to the more complex issues.

## The Background Theory: Kripke-Style Externalism

Saul Kripke's series of lectures published as *Naming and Necessity* (Kripke 1980) outlines what has now become one of the leading theories of how language connects to the world. Similar and complementary views were developed at more or less the same time by Hilary Putnam (1975) and Tyler Burge (1979). The distinctive feature of these views is a form of externalism: what grounds meaning and determines what sentences and expressions are about is external to the speaker's mind. If you look at just the speaker of 'John is in Paris', you won't find out what the expression 'Paris' is about. Moreover, it is not determined 'computationally': reference determination is not about how a particular symbol computationally integrates into the neural net calculations. It's not computational in the sense of being about the internal computational structure of the representational item itself. It's all about external relations to the world—looking at internal computation structure is just wrong-headed through and through. What determines that an utterance of 'Paris' is about Paris has to do with the history of use of that name. Here is how Kripke introduces his basic idea:

> Someone, let's say, a baby, is born; his parents call him by a certain name. They talk about him to their friends. Other people meet him. Through various sorts of talk the name is spread from link to link as if by a chain. A speaker who is on the far end of this chain, who has heard about, say Richard Feynman, in the marketplace or else-where, may be referring to Richard Feynman even though he can't remember from whom he first heard of Feynman or from whom he ever heard of Feynman. He knows that Feynman is a famous physicist. A certain passage of communication reaching ultimately to the man himself does reach the speaker. He then is referring to Feynman even though he can't identify him uniquely.
>
> (Kripke 1980: 91)

---

[1]  It might seem surprising that we use a variety of externalist theories that are often seen as competitors. It's not strange if, as we do, you endorse a form of metasemantic pluralism: There are many metasemantic mechanisms (and there could be even more than we currently know of).

Note that on this view, a speaker can use a name to talk about something or someone even if that speaker has no ability to describe that thing correctly. It is not the speaker's beliefs about what 'Paris' denotes that determines what she denotes by 'Paris'.

Aboutness is determined by an external chain of communication. Here is Kripke's rough summary of his view:

> An initial 'baptism' takes place. Here the object may be named by ostension [ . . . ]. When the name is 'passed from link to link', the receiver of the name must, I think, intend when he learns it to use it with the same reference as the man from whom he heard it. If I hear the name 'Napoleon' and decide it would be a nice name for my pet aardvark, I do not satisfy this condition.    (Kripke 1980: 96)

The structure of such a theory is relatively simple. It has three parts:

1. There's an **introductory, anchoring, event**, where an expression is 'hooked up' to some part of the world ('Paris' to Paris, 'Napoleon' to Napoleon, 'zebra' to zebras, 'chair' to chairs, etc.). Kripke suggested this could happen through a baptism (as in the example in the quotation above) or by a description being used to pick out the thing talked about (if we said, 'let "Alfred" be the name of the first person to buy a copy of this book').

2. Then there's **a chain of transmission** from person to person. This is what Kripke also calls a communicative chain. Kripke stipulates that this chain has to be reference preserving (more on that below).

3. Then there's a speaker using the expression at some point in the chain: She can use, say, 'zebra' to talk about zebras **by virtue of being part of a communicative chain** that started with zebras (e.g. that started with a baptism where a speaker said: call those kinds of animals 'zebras'.)

There are two notable and relevant features of this view:

1. Even if you knew everything that had ever happened to the speaker, you would not know what she is referring to (what she is talking about). That is determined by historical facts that are independent of that particular speaker (the beginning of the communicative chain, which could have started before the speaker was born). In particular: if you look inside the head of the speaker, there's no fact 'in there' that will tell you what she is talking about.
2. The speaker could be radically wrong about what she is talking about—she could be wrong about what her use of 'Napoleon' refers to (she need not know what she is talking about because she need not know what is at the origin of the communicative chain).

Of course, often things are more complicated and messy. And the Kripkean view has inspired mountains of theorizing, both defending and furthering his externalist view (as in Salmon 1986, Soames 2002 and more recently, in a textbook presentation: our 2018) and responding to it (as in the causal descriptivism of Lewis 1984 and Kroon 1987, or the two-dimensional semantics defended in works like Chalmers 2006).

While important, we don't think that we need to engage with this literature too much here—our aim is to get as far as we can with (and simply assuming the truth of) the key Kripkean insight and picture of reference. With that mind, let's consider how this can help us understand content attribution to ML systems.

## Starting Thought: SmartCredit Expresses High Risk Contents Because of its Causal History

The externalist story we just outlined has two ways of describing the anchoring of the content of ML systems:

1. SmartCredit's history includes an anchoring event, anchoring on high risk.
2. SmartCredit is a link in a transmission chain leading back to high risk.

The obvious problem with this line of thought is that there's no simple way to apply Kripke's picture directly to SmartCredit. There's just nothing that looks like a standard anchoring event in SmartCredit's history. SmartCredit never points to anything (let alone to the property of high risk), it never descriptively singles out anything, it never has referential intentions. And SmartCredit can't be inheriting a semantic connection of high risk from elsewhere (from the programmers, perhaps) in the usual way, because there's nothing that looks like the standard transmissive link in SmartCredit's case. SmartCredit has no intention to use a term in the same way as those from whom it received the term. In sum:

- There is nothing like a Kripke-style baptism event.
- There is no intention to refer to high risk, no pointing at high risk, no descriptive identification of high risk.
- And there is nothing like Kripke-style proper transmission. In particular, there are no reference-preserving intentions.

So while—for reasons we spelled out above—externalism might seem plausible at first glance, the versions that are available off the shelf seem initially unpromising.

## Anthropocentric Abstraction of 'Anchoring'

The problems above take this form: they point out disanalogies between the ways in which humans initiate and participate in communicative chains and the ways AI, on our proposal, would do so. There will obviously be very many such differences. Humans are animals that engage with the world in ways computers can't. We have all kinds of inter- and intra-personal experiences that computers lack. To be open to the idea that systems very different from humans can have content, we need to engage in what we call 'anthropocentric abstraction': the effort to find some more abstract description of the structure that leads to content attribution for humans—a description that moves away from the contingencies and limitations of our peculiarities.

There's a general structure to that process of anthropocentric abstraction:

- We start with a pattern instantiated by human peculiarities.
- There is then a hierarchy of degrees of abstractions: what we are looking for is a degree of abstraction that preserves sufficiently many important features of the original phenomenon. It's not too abstract and it's not too focused on specific details. Call that **'the abstractive sweet-spot'**.

This process of abstraction can be done bottom-up (starting with particular cases and working one's way up) or top-down (starting with a general theory and working one's way down to specific cases). Kripke's strategy is the former: he starts with cases that we all agree are instances of reference, and then he builds a very thin theoretical framework on top of that. What he in effect does is give a brief sketch of how we humans typically do it, based on reflection on a few cases. He does not start by an a priori articulation of a general condition that has to be imposed on reference and then looking for human behaviour that satisfies those conditions. We will use the Kripkean bottom-up approach. We start with the assumption that an ML system is, say, a high risk detector. We then explore its history and we ask: what in that history corresponds to what we find in the human case? Does any of it match—at an appropriately abstract level—some of the components of what Kripke finds in the human case?

## Schematic AI-Suitable Kripke-Style Metasemantics

In order to de-anthropocentrize the Kripkean story that the analogue of anchoring for AI systems is to be found in their neural net training, very roughly, our proposal is this:

> SmartCredit's outputs express the property of high risk because the training of SmartCredit's neural network was done against the property of high risk, thereby anchoring the program to that property.

This is the coarse-grained answer, but the details will matter a great deal and they are in large part unsettled. Some of the relevant details involve how neural nets are trained:

- A generic initial neural net is given samples from a large pool of training cases.
- Each training case has been hand-coded ('high risk' versus 'low risk'), for example by the programmers.
- The AI's output for the training case is then compared to the hand coding using some scoring function evaluating how well the AI classified the training case.
- That score is then used to update the weightings of the node connections in the neural net.

Our suggestion is that SmartCredit's outputs express the property of high risk because SmartCredit was given training cases that were hand-coded for being high risk lendees or low risk lendees, and was then scored highly for categorizing cases hand-coded high risk as high risk and scored poorly for categorizing cases hand-coded high risk as low risk. Its net is then adjusted on the basis of that score. After some (indefinite) number of iterations of this process, SmartCredit's outputs gain representational content.

Here, in slogan form, is the proposal:

**AI Anchoring**: SmartCredit is anchored in high risk via a scoring function that scores well for matching high risk hand coding and low for not matching.

We think that this is a plausible starting point, but it's just a schematic view right now, because there are a number of choice points that we'll encounter as we think through the details of hand coding, of scoring functions, and of updating procedures.

## Complications and Choice Points

**Hand coding choice points**: We've been setting out a Kripke-inspired picture on which SmartCredit is anchored in the property of high risk because it's trained on a bunch of cases that have been hand-coded as high risk or low risk. But there's more than one thing we might mean by 'hand-coded as high risk or low risk'. To see this, consider two cases:

(C1) SmartCredit's training set was assembled by programmer Pat. Pat has gone back through old bank records, found numerous instances of people who did and did not default on their loans, and then put together files of the input data (at the time of loan application) for these people, together with a label of 'high risk' for the actual defaulters and 'low risk' for the actual non-defaulters. But Pat makes a few mistakes along the way. Among the thousands of test cases, there are three (for three individuals A, B, and C) that Pat marks with a 'high risk' label even though they didn't, in fact, default on their loans. When SmartCredit is trained on this data set, is SmartCredit being trained on a data set hand-coded for the property of being high risk, or on a data set hand-coded for the property of being high risk or one of A, B, and C?

(C2) Pat wanted more cases than were available in the bank's lending history, so created a number of additional fictional cases. Pat uses the best of her financial knowledge to design fictional cases of defaulters and non-defaulters, and then creates initial data sets suitable for those fictional cases, and labels those cases with 'high risk' and 'low risk' labels as appropriate.

But, of course, there is no independent fact of the matter of whether these cases are genuinely high risk or low risk cases. When SmartCredit is trained on this data set, is SmartCredit being trained on a data set hand-coded for the property of being high risk, or on a data set hand-coded for the property P of being someone who Pat thinks would be high risk?

We can now consider various particular versions of the general Kripke-inspired metasemantics:

(K1) SmartCredit is anchored to a property P if P is the property in fact shared by all the training cases hand-coded with the same label.

(K2) SmartCredit is anchored to a property P if P is the property that the hand-coder intends to be indicating by marking training cases with a given label.

Consider hand coding. Suppose that some of the training cases are mislabelled in the hand coding—cases that are in fact high risk lenders are marked as low risk lenders. How will this affect what property the neural net is anchored in? (The *intended* property? Some disjunctive property?) Or suppose we don't use actual cases as training cases, but fictional cases, so that there is no independent fact of the matter about how the cases are *correctly* hand-coded. What effect will such training cases have on content fixation?

**Scoring choice points**: When SmartCredit is being trained—and thus, on our Kripke-inspired picture, hopefully being anchored to the property of high risk—its outputs for a test set are compared to hand-coded evaluations of the test set. We then want to give SmartCredit feedback based on how well it did at categorizing the test cases. But there are many notions of 'how well' that could be used here.

Consider a complication that we've been sweeping under the rug. SmartCredit, like many AI classifiers, doesn't produce *binary* classification judgements. When given Lucie's data, it doesn't just report that Lucie is high risk or report that Lucie is low risk. Instead, it assigns probabilities that Lucie is in each category. So SmartCredit might report that Lucie is 0.8 likely to be high risk and 0.2 likely to be low risk. Now suppose that SmartCredit produces probabilistic outputs like this for thousands of cases. For each of these cases, we also have hand-coded evaluations of whether the person is genuinely high or low risk. Now we want to assess how well SmartCredit did. That can't just be a count of how many cases SmartCredit got right and how many SmartCredit got wrong—the 'how well' assessment needs to take SmartCredit's probabilities into account.

What we need is a *scoring function*. But there are many ways to design a plausible scoring function, and different scoring functions are in fact used in different AI applications. Let's consider briefly two scoring functions. One is the **Brier score**. To obtain SmartCredit's Brier score, for each case we take the difference between SmartCredit's assigned probability of being high risk and the actual 'probability of being high risk' (1 if the case is hand-coded as high risk; 0 if it is hand-coded as low risk). We then square each of those differences and add them. So for a test set S, SmartCredit's Brier score is:

$$1/|S| * (\sum_{(i \text{ in } S)} \left( L_i - A_i \right)^2)$$

where $L_i$ is SmartCredit's probability that the ith case is high risk and $A_i$ is the actual probability that the ith case is high risk.

Lower Brier scores indicate better accuracy; higher Brier scores indicate worse accuracy.

Another scoring function is the **log-loss score**. To get SmartCredit's log-loss score, take SmartCredit's assigned probability that a given case is high risk, and then take either the logarithm of that probability (if the case is hand-coded as high risk) or the logarithm of 1 minus that probability (if the case is hand-coded as low risk). So for a test set S, SmartCredit's log-loss score is:

$$-1/|S| * (\sum_{(\text{i in S})} \left( A_i * \log\left( L_i \right) + \left( 1 - A_i \right) * \log\left( 1 - L_i \right) \right)$$

Again, lower log-loss scores indicate better accuracy.

Brier scores and log-loss scores won't in general be the same, and so training a program to minimize the Brier score won't in general produce the same behaviour as training a program to minimize the log-loss score. For example, the log-loss score punishes large probability errors more severely than does the Brier score. Consider the accuracy penalties, for both scoring functions, of outputting probabilities of either 0.01 or 0.001 for a case that is in fact high risk:

Brier score:
Output = 0.01: Brier score = $(1-0.01)^2 + (0-0.99)^2 = 1.9602$
Output = 0.001: Brier score = $(1-.001)^2 + (0-0.999)^2 = 1.996002$
Log-loss score:
Output = 0.01: Log-loss score = $-\log(0.01) = 2$
Output = 0.001: Log-loss score = $-\log(0.001) = 3$

The Brier score increases the accuracy penalty for the second case by less than 2 per cent, while the log-loss score increases the accuracy penalty for that case by 50 per cent. (Brier scores for individual cases are bounded at 2 while log-loss scores are unbounded, so in extreme cases the penalty increase for the Brier score tends to 0 while the penalty increase for the log-loss score increases arbitrarily.) So an AI system trained using a log-loss scoring function is, compared to a system trained using a Brier score, made more likely to avoid extreme probability errors.

Of course, if SmartCredit is getting everything right, it doesn't matter which scoring function is used. But no AI system is going to get every judgement right. Just for a toy case, let's suppose that the financial prospects of bitcoin speculators are particularly difficult for SmartCredit to evaluate. (For whatever reason, the kinds of correlations between social media footprint and creditworthiness that SmartCredit relies on are much less robust among bitcoin speculators than among the general population.) So SmartCredit's assigned probabilities for bitcoin speculator test cases tend to produce extreme errors—SmartCredit is often highly confident that a genuinely risky bitcoin speculator is low risk, or vice versa.

Now consider two properties to which SmartCredit might be anchored, Kripke-style: (i) being high risk, or (ii) being a bitcoin speculator or a high risk non-bitcoin-speculator. There could then be two different elaborations of the Kripke-inspired picture, which predict anchoring onto these different properties:

(K1) SmartCredit is anchored to property P if SmartCredit is trained using a training set hand-coded for some property Q such that P is the simplest property that produces a reliably low Brier score compared to the hand-coded Q facts.

(K2) SmartCredit is anchored to property P if SmartCredit is trained using a training set hand-coded for some property Q such that P is the simplest property that produces a reliably low log-loss score compared to the hand-coded Q facts.

Then suppose that according to the metasemantic view (K1), SmartCredit represents being high risk, while according to the metasemantic view (K2), SmartCredit represents being a bitcoin speculator or a high risk non-bitcoin-speculator. Both (K1) and (K2) are particular ways of filling out the general externalist Kripke-inspired metasemantics—how could we decide which of (K1) and (K2) is the right way to abstract a non-anthropocentric metasemantics from the Kripkean starting point?

At this point, we turn to the meta-metasemantics. Given our interpreter's knowledge-maximization picture of the meta-metasemantics, we need to know whether a (K1)-style metasemantics or a (K2)-style metasemantics for SmartCredit will maximize the knowledge we obtain through our interactions with SmartCredit. The answer to that question is then sensitive to a number of externalist features of the social and environmental setting in which we use SmartCredit. For example:

- If the environment is heavily populated with bitcoin speculators, (K1) will have SmartCredit inaccurately labelling them as (e.g.) high risk, since the Brier score doesn't weight the extreme probability errors for these cases heavily enough to influence the property tracked, while (K2) will have SmartCredit accurately labelling them as either bitcoin speculators or as high risk non-bitcoin-speculators. (K2) would then, to that extent, be more conducive to interpreter knowledge. But if the environment is sparsely populated with bitcoin speculators, the stronger property represented by SmartCredit according to (K1) might lead to more knowledge on our part (since we can also infer the disjunctive property).

- What we plan to do with the classification we get from SmartCredit can influence which property is knowledge-maximizing for us. Suppose the bank has a policy of not lending to bitcoin speculators, and the cases of both Simon the bitcoin speculator and Lucie the non-bitcoin-speculator are both given to SmartCredit. We then form both (i) classificatory beliefs about Simon and Lucie, and (ii) a secondary practical belief about how we ought to treat Lucie (that we should or should not give her a loan). When the content of the secondary practical beliefs rely on the SmartCredit content ascribed by metasemantics (K1), our secondary belief about Lucie is knowledge (because *bad risk* is a good reason to deny a loan). But when the content of the secondary practical beliefs rely on the SmartCredit content ascribed by metasemantic (K2), our secondary belief about Lucie is not knowledge (because *bad risk or bitcoin speculator* is not a good reason to deny a loan). So (K1) has some knowledge maximization effect over (K2). But if the bank has no policy against loaning to bitcoin speculators, the secondary practical question arises for Simon as well. (K1) and (K2) make that secondary belief about Simon not knowledge, but (K2), and not (K1), makes the classificatory belief about Simon knowledge. So in this environment, (K2) has some knowledge maximization effect over (K1). In general, since log-loss scoring functions avoid *extreme* errors more than Brier scoring functions, (K2) will be a better knowledge-maximizer than (K1) in cases in which our subsequent use of the categorizing is in a context in which we care a lot about avoiding bad errors. So, for example, a 'guilt-innocence' detector might be more likely to be knowledge-maximizing when it's guilt detecting according to the (K2) log-loss metasemantics than when it's guilt detecting according to the (K1) Brier score metasemantics, given the nature of the other beliefs we're going to form based on the guilt-innocence categorization.

**Update choice points**: Even more pressingly, there are many ways of going from a scoring of the AI output to a specific alteration of the neural net node connection weightings (many complicated papers are written about this in the AI literature). Clearly not all ways of updating are going to produce the same content (intuitively, 'inverting' the update function for SmartCredit's training should 'invert' its representational contents), so again there's room for interaction between the details of the update function and the details of the anchoring.

## Taking Stock

Here is what we have done:

We started with an outline of Kripke's causal chain metasemantics.

We observed that the details of this metasemantics aren't straightforwardly applicable to AI systems.

We suggest that the Kripkean metasemantics is an anthropocentric instance of a larger class of metasemantic principles.

We took some initial steps toward de-anthropocentrizing, proposing an AI-friendly version of anchoring.

Finally, we outlined some choice points for that theory.

## Appendix to Chapter 5: More on Reference Preservation in ML Systems

We just expressed optimism about anthropocentric abstraction of 'reference-preserving intentions'. We should add that a full theory will have to engage with a range of interesting differences between humans and ML systems. There are fundamental differences between programs and people in the way that information is transmitted and this will matter to whether reference chains are being preserved in 'the right way'. Here are some additional cases to consider:

(1)  Suppose we have programmed a neural network on a particular computer in Oslo. That network then gets trained on lots of duck photographs. Let's assume that's enough for anchoring, and as a result, that neural network's outputs are now about ducks. We then email that program to another computer in Austin. On that computer in Austin there's now a new token-distinct but type-identical program. Is that program part of the same referential chain? Are its states also about ducks?

(2) We can easily imagine more complicated cases. Suppose we have a neural network that's been trained to recognize photographs of Pacific black ducks. We want to make a new program that recognizes photographs of eider ducks. Rather than retrain a new neural network on a new collection of eider duck photographs, we take the neural network weightings of the Pacific black duck recognizer together with a description of the typical colouring of an eider duck and apply a metaprogram that reweights a neural network to transform its recognitional sensitivities. We end up with a new program that functions well: it reliably (but not always) labels eider duck photographs as hits and non-eider duck photographs as misses. But are its reports about eider ducks? That depends on whether this more complicated method of causal transmission counts as reference preserving. More generally, programs offer opportunities for causal transmission and manipulation (by human programmers, by other programs, and so on) that aren't available with people, and a good non-anthropocentric version of externalism needs to include tools for deciding which of these opportunities are reference preserving and which are not.

(3) Suppose we are trying to do early cancer detection, so we create a machine learning cancer detector. We train it in the usual ways, giving it a sample set of cases and a scoring system on those cases. But once the program has been trained up, we also allow it over time to use data mining methods to look for additional patterns in the new cases and dynamically adjust its own categories. That means that over time, the program might end up categorizing in ways that largely disagree with the scoring on the original training set. But we can imagine multiple ways in which this change in categorization behaviour might go. We might discover that the program has become a better cancer detector than we were—that we had made mistakes on some of the original training set, but that the program is now able to detect cancer better than we could, and is correcting those mistakes. Or we might discover that the program has become a deeper characterizer than we were. Perhaps we learn that 'cancer' is actually a confused category, one which lumps together medically distinct conditions and artificially separates other conditions that are medically similar. The program as it develops has got onto a different, more medically robust category, and is tracking that rather than cancer. Or we might discover that the program has gone off the rails entirely—that its dynamic adjustment of its own categories has drifted hopelessly away from anything that we ever wanted to track, and that it's now just tracking some random and medically uninteresting collection of blood chemistry features. In each of these cases, we're faced with the question of whether the program's outputs are still about cancer or have come to be about new categories. Answering that question, from an externalist perspective, requires determining whether the dynamic development of the program is properly reference preserving. No simple application of the Kripkean model is going to answer that question.[2]

---

[2] It is unclear how different this is from human cases: we can imagine a human researcher, who starts off as a straightforward cancer researcher, whose research develops in each of the three ways sketched above, but who keeps using the word 'cancer'. We're then confronted with similar questions about whether his use of the word 'cancer' is still part of the causal chain to which he was originally introduced.

(4) Lockdown is a public safety AI system, designed to assess the risks of venturing outdoors. Lockdown takes a wide variety of input data on weather, crime, epidemiology, economic markers, social media activity, and so on, and delivers a verdict of 'safe' or 'unsafe'. But Lockdown delivers location-specific recommendations. Albert, running Lockdown in Oslo, gets an output of 'safe', meaning that it is safe to go outdoors in Oslo. Beth, running Lockdown in Stockholm, gets an output of 'unsafe', meaning that it is dangerous to go outdoors in Stockholm. Lockdown's outputs are thus context-sensitive—a Lockdown output of 'safe' means, roughly, that it is safe *here*, where 'here' picks out the place being evaluated on that run of Lockdown.

There are two interrelated problems about representational content that are raised by an AI system like Lockdown. First, if Lockdown's contents are best understood as context-sensitive, stating that things are safe or unsafe in the context of utterance, what counts as the context of utterance? Second, what determines that these kinds of AI outputs are context-sensitive, rather than context-insensitive? How does an appropriately de-anthropocentrized metasemantic story predict which AI outputs are context-sensitive and which are not?

These four cases illustrate the kinds of complexity that will arise in developing a complete externalism for ML systems. One possibility here that we find quite plausible is this: many of these questions do not have predetermined answers. What will count as a correct answer in many of these cases will depend on **decisions** we as speakers (and as speech communities) make as the engagement with ML systems become more entrenched. Maybe in 100 years, we will have developed stable patterns of how to interpret these kinds of cases.

# APPLICATION

## *Names and the Mental Files Framework*

### Does SmartCredit Use Names?

In the previous chapter, we outlined a proposal for how a de-anthropocentrized version of Kripke's theory could explain how SmartCredit could represent the property of being high risk. In the sentence we started with 'Lucie is high risk', that property is attributed to a person, Lucie. We now turn to the question of how SmartCredit can refer to Lucie.

An initial observation: we cannot assume that the lexical item 'Lucie' plays a significant role in SmartCredit's neural network. What happens is this: some information about Lucie is initially fed into the system. This will include various financial and demographic information. The system will then 'collect' more information. If we idealize a bit, we can think of the system as ending up with a potentially enormous collection, C, of information. What SmartCredit is then programmed to do is assess whether an individual that has all the properties in C is high or low risk. In the case we have been considering, the conclusion is that the individual is high risk. If this is the right description, SmartCredit is different from regular English speakers in that proper names play a rather limited role in its computational structure. Property clusters play a central role. What we assume when we translate the output into the sentence: 'Lucie is high risk' is that Lucie satisfies a certain property cluster. We can think of the output as in effect being of the form:

Something satisfying C is high risk.

Then there's a background assumption, that a particular person, Lucie, satisfies C. What happens at the output point is that this assumption is added and we present the output as if it's directly about Lucie, i.e. as:

Lucie is high risk.

If the story we've just told is correct, that presentation of the output is tendentious because it relies on an implicit assumption: that the person we refer to with the name 'Lucie' is someone who satisfies C.

One advantage of this picture is that, if correct, it means that we don't need to add an account of how SmartCredit can have competency with names. What we need is an account of how it can use the predicates that are components of C (and we gave an account of that in the previous chapter), and then an account of predication (which we will give in the next chapter).

A disturbing feature of this account is that the real output of SmartCredit, i.e. *that something satisfying C is high risk*, could become impossible to understand and track. The property cluster that C is an abbreviation for will potentially track properties and interconnections between properties that we cannot express. Initially, the input we give SmartCredit might be tractable for us, but as it starts collecting information from varied sources, the resulting complexity will, in many cases, be too complex for a human mind, even if we did have the terminology to express it.

If the output is something we cannot grasp, that makes the assumption that Lucie satisfies the conditions in C one we cannot fully grasp (or understand). If we can't grasp it, we can't assess it. In other words, we are then building in a tacit assumption that we're incapable of assessing. Since C will contain an enormous amount of information, we can safely assume that it will frequently happen that some of the information doesn't apply to the individual to whom we take it to apply, in this case Lucie. We then face questions about how to treat the output of SmartCredit when some, but not all or most, of the information in C fails to apply to the individual we interpret the output to be about.

In other words, if SmartCredit has no capacity for representing in some way analogous to how we represent using proper names, then we face both communicative and epistemic obstacles in our engagement with it.

## The Mental Files Framework to the Rescue?

We just outlined some problems for a certain model of how AI represents Lucie. Anyone familiar with the last 120 years of philosophy of language will recognize that analogues of these problems have been extensively discussed. One of the fundamental divides in the literature is between theories according to which names are clusters of descriptions and theories according to which they are not. In the above paragraph, we in effect first proposed a descriptivist view and then raised some problems for it.

Rather than rehearse that entire debate here, we will do as we did in the section of predicates: use one of the standard theories and see if it can be applied to an AI system like SmartCredit. The framework we will appeal to is the so-called mental file framework. Early work on this includes Lockwood (1971), Perry (1980), and Evans (1982). Recanati (2012) is a recent and comprehensive presentation and defence of the view. In what follows, we will rely in large part on Recanati's view.

At the core of the theory is the idea that human cognition centrally involves clusters of properties. In saying that they are clusters, we mean that the properties are presented as co-instantiated. Such clusters are, in this literature, called mental files. Using the metaphor of files, we can talk of the files as consisting of properties, where that is to say that these properties are co-instantiated. Here is Aiden Gray's useful summary of the view:

> The role of a file is to collect and store information derived from a single object. Files are temporally enduring—an agent maintains a file over time, adding new information derived from the same object. A thinker can employ a file to think and reason about the referent of the file.   (Gray 2016: 348)

Three important observations about the properties contained in a file:

1. **A negative thesis**: The object the file is about is not the object that has all the properties in the file. In Evans' terminology: the denotation of a file is not achieved through fit.

2. **A positive thesis**: Reference to an object is determined by various external relations. We'll say more about these below; they can be epistemic, historical, and causal.

3. **A corollary**: The file for a person, say Lucie, can fail to describe her: As long as the external connection holds between Lucie and the file, the file is about Lucie even if *all* the information contained in it fails to apply to Lucie.

According to the mental file framework, we should think of proper names as associated with mental files. For each name, a different mental file. The files can evolve over time (as we gain and lose information), can be combined, and can sometimes be divided.

Two important questions arise for this kind of view. First, what does it take for a mental file to be associated with a name: e.g. under what conditions is a file the Lucie-file? The leading theorists are quite vague on this. Again, Gray gives a good description:

> One sometimes sees the claim that a particular file is 'labeled' with a name—for example, Recanati (2012, pg. 190) and Lockwood (1971, pg. 208). This is, at best, a placeholder for an account of the connection between a file and a name, and, at worst, a misleading metaphor.

We think Gray is right that the mental file framework is deeply metaphorical and that these metaphors are both integral to the attractiveness of the view and potentially misleading. Our heads contain no real files, no labels, and no filing cabinets. Insofar as the theory trades heavily on these metaphors, it's in danger of misleading us. But for now, we'll put those concerns aside. We'll focus on the good parts of the theory and use them to help us understand SmartCredit.

The second question that is important for our purposes is how to understand the relation between a file and its referent. We know from 1 and 2 above that it's not through fit, but through some kind of external relation. To assess the theory, we will need more details. Recanati's favoured term for the relation is 'epistemically rewarding'. He says:

> The characteristic feature of the relations on which mental files are based, and which determine their reference, is that they are epistemically rewarding […] They enable the subject to gain information from the objects to which he stands in these relations.
>
> (Recanati 2012: 35)

Perception of an object is a paradigm of an epistemically rewarding relation. That, however, obviously cannot be the full story about names because we can talk about Cicero using 'Cicero' despite never smelling or touching or hearing or seeing him. So if we use 'Cicero' to talk about Cicero in virtue of having a file that stands in an epistemically rewarding relation to Cicero, then such relations must include, for example, information gained through testimony.

## Epistemically Rewarding Relations for Neural Networks?

To extend Recanati's framework AI, we need to de-anthropocentrize the notion of an epistemically rewarding relation. We have ideas about what would constitute such relations for humans and, as we have seen, perception is often presented as a paradigm. Independently of considerations having to do with AI, we know that this is too anthropocentric: surely there can be creatures that refer using, say, 'Lucie' to refer to Lucie, just as we do, but don't perceive the world the way we do or, more generally, gain knowledge about the world in ways completely different from us. Maybe they rely entirely on telepathy. Maybe they gain knowledge in ways we have never thought of or cannot fully understand. It would be parochial to stipulate that such creatures cannot use 'Lucie' to refer to Lucie.

The kinds of AI that we are discussing are, to a significant extent, alien. In particular, the algorithms governing neural networks will do things we don't or can't fully understand. As a result, the epistemically rewarding relation will look different from the human paradigms.

What we need to do is familiar by now: the models developed by mental file theorists must be de-anthropocentrized and applied to AI systems like SmartCredit. The resulting theory should satisfy the two core elements of our positive theory:

- It will be externalist (because the relation between a file and the object the file is about is external to the speaker).
- It will be developed by abstracting away from anthropocentric components of existing theories.

Note that when Recanati and other mental file theorists describe epistemically rewarding relations that human speakers rely on, they do so in rather abstract terms. They don't, for example, go into the details of how particular perceptual systems like smell or touch work (and end up being epistemically rewarding). They are not even particularly clear on just what counts as 'rewarding' or 'epistemic'. It's just assumed that, say, perception is a paradigm of something that's an epistemically rewarding relation. Similarly, it is not to be expected that our theory goes into great detail of all the various epistemically rewarding and reference determining relations that AIs stand in to objects. What we have to say here will be at a fairly high level of abstraction.

The first step is to appeal to the meta-metasemantic principle of knowledge maximization from Chapter 4. According to this principle, the relevant epistemically rewarding relation should be knowledge maximizing. The most natural way to do that is to build knowledge maximization directly into the account of 'epistemically rewarding'. Gray cites the following passage from Williamson:

> A causal relation to an object (property, relation,…) is a channel for reference to it only if it is a channel for the acquisition of knowledge about the object (property, relation).
>
> (Williamson 2007, 264, ellipses in original)

The more detailed story then is an answer to the following question: which specific such relation maximizes the interpreter's knowledge if used to determine reference? The answer to this will to a large extent vary between AI systems. It will depend on the details of how the system works, how it was created, and how it is used.

In saying this, we don't mean to be defeatist about a general theory. Some of the material from the previous chapter is directly relevant here. In Chapter 5, we developed an account of anchoring for predicates that relied on the idea that a property anchors *a training process*. In many cases, elements of the training process will also be important in understanding the epistemically rewarding relation to an object that determines reference of singular terms. Here it is natural to appeal to some of the central notions developed by Gareth Evans (one of the early proponents of mental file theory). According to Evans, the denotation of the use of a term T is fixed by what he called the 'dominant source' of the information associated with T. This was the way Evans developed Kripke's causal theory. He agreed with Kripke that reference is not fixed by descriptive fit, i.e. not fixed by what the associated descriptions pick out. However, he diverged from Kripke in giving the associated descriptions a reference-determining role: not through fit, but through an external causal relation. The notion of a dominant causal source is now at the centre of the theory and will need further elaboration. However, for our purposes, we will simply use this schematic idea and apply it to AI. The schematic idea is this:

> *Denotation through dominant causal source:* A system can denote an object that is the dominant causal source of a set of information given as input in the training stage.

If we return to our simple case of SmartCredit and Lucie, the initial stage involves information being fed into the system. Call the conjunction of that, C. Think of C as a mental file. There will be an object that is the dominant causal source of C. If things go well, this will be Lucie. If so, SmartCredit can refer to Lucie through Lucie being the dominant causal source of C. If so, it is correct to describe the output of the system as being of the form: *Lucie is high risk* (and not just: 'Someone satisfying C is high risk').

## Case Studies, Complications, and Reference Shifts

There are several concerns about Evans' theory. Central among these is that we need more clarity in what counts as a dominant source of a body of information. That it is dominance that determines reference means that not all of the associated information needs to have the referent as a source: there can be misinformation mixed in that doesn't have the referent as a source. Evans is also clear that dominance is not simply a matter of quantity: it is not a matter of a simplistic counting information and then locating the source of the majority. Some of the information is more heavily weighted than others. Evans is also clear that over time, dominance can shift. He illustrates this with the example of 'Turnip' (1973: 306). The example involves a youth, A, with the nickname 'Turnip'. He leaves his village at an early age. Many years later, a different person, B, settles in the village. The old villagers believe that A has returned and refers to B using 'Turnip'. At first, the dominant causal source associated with 'Turnip' will be the A. Over time, however, this can change: the dominant source of information can be shifted from A to B. In Evans' example, it's easy to see how that can happen: as villagers see more and more of B, the file associated with 'Turnip' will gradually fill up with more significant information that has B as its source. The information that has A has its source will gradually fade into relative insignificance. The possibility of a name shifting referent over time was one of the central motivations for Evans' theory. Evans argued that Kripke's theory couldn't account for such reference shifts and that his alternative could do so easily.

We mention this because the kind of flexibility that Evans' version of the mental file theory provides can be useful for interpreting AI. First, it should be possible for the initial data to contain what we would naturally classify as *misinformation* about A. One way this could happen: some of the descriptive material, D, in the A-file has another object, B, as its source. It correctly characterizes

B, but fails to describe A. As long as B isn't the dominant source of the information in the file, the file as a whole can have A as its denotation. It simply contains the misinformation that A is D. Moreover, gradual shifts can happen as follows: the information we feed the system can initially have Lucie as its dominant source. As information is added over time, the dominant source can shift to another person. This can happen in two ways:

(i)  What counts as dominant information can shift even if the total amount of information doesn't shift. This can happen even as the informational content of a file in an AI system is stable.

(ii) Information can be added to the file. This can change the dominant causal source and that again can result in reference shift for the file as a whole.

Whether (i) is an option will depend on how dominance is understood. As we see it, dominance is unlikely to be understood independently of human interests. What counts as dominance is not an objective feature that can be read off the world independently of what interpreters care about. If that is right, then a gradual change in what we care about can result in a change in what the file refers to. Note that these issues about how to understand dominance comprise a meta-metasemantic question. Our guiding principle is knowledge-maximization and it tells us that dominance should be construed in a knowledge-maximizing way.

Many cases will not fit cleanly into either of (i) or (ii) above. Consider the following: suppose we have an AI system that is set up to make assessments of economic health. Initially, it is fed information about the economic situation in the US. It starts giving outputs of the form 'The economic outlook is G'. Since the US is the dominant source of the information the system is operating on, we should interpret this to mean that the economic outlook in the US is G. Doing so is knowledge-maximizing. Now suppose that over time, the system starts to focus more specifically on the input of economic data from California (it is still using the entire data set, but changes its focus to California). This could happen for several reasons: maybe that turns out to be the data that's most predictive or that its algorithms can do the most with. It keeps producing outputs of the form 'Economy is doing great' or 'Economy is doing poorly'. Now we ask: *which* economy does it refer to, the US, or just California, or something else?

The general form of an answer to that is guided by the general principle that it refers to whatever place it's in an epistemically rewarding relation to. Our knowledge-maximizing principle tells us to construe 'epistemically rewarding' as a relation that is knowledge maximizing for us as interpreters/users. However, just saying that leaves us with a range of possible outcomes. Here is one possible outcome: it might turn out that its predictions are more accurate about California, although it does well enough for the US as a whole. If so, there are competing considerations for how to understand 'maximizing'. It will matter what we care the most about: precise knowledge or maximally general knowledge? The answer to that, again, might depend on what we are going to do with this information and how we use it to generate further knowledge. This case shows how the metasemantics, guided by knowledge-maximization, will be sensitive to our interests and activities in ways that are hard to predict.

Next consider an AI system used by law enforcement to help determine who is guilty of various crimes. It's in effect an AI-detective and it tells about the degree of guilt of various subjects. Initially, we feed it information about Lucie's activities. After processing these, it outputs 'Innocent', if it finds no violations of any laws. If Lucie has had a couple of speeding tickets, the output is 'Guilty of minor traffic violations'. The referential issue is fairly simple since Lucie remains the dominant source of information throughout. Here is a more complicated case: we start to feed it data about a supposed mob family. The data includes a broad range of actions performed by several people over a period of time. This makes the referential question more difficult to settle. We need to decide whether it is tracking guilt of the *organization*, or guilt of *specific people in the organization*. After all, if the organization is guilty, that is the result of various crimes committed by the members of the crime family. In tracking the family, it is tracking the individuals. We can use the knowledge-maximization principle to help us adjudicate: is the epistemically rewarding relation, i.e. the knowledge-maximizing relation, one that leads to the family or to particular individuals? As in the previous case, that might depend on our interests and what we end up doing with that knowledge. It could, for example, depend on how we use it to generate new knowledge.

What both these cases are meant to illustrate is that the correct externalist interpretations can be sensitive to variations in our epistemic goals and practices. This follows quite naturally from the general principle of aiming for a metasemantics that maximizes knowledge. The measure of maximization can vary in various ways and the two cases illustrate that.

## Taking Stock

Here is what we have done:

1. We started with an outline of the mental file metasemantic framework.
2. We observed that the details of this metasemantics aren't straightforwardly applicable to AI systems—in particular, we need to abstract on the notion of an epistemically rewarding relation.
3. We took some initial steps toward de-anthropocentrizing, proposing an AI-friendly version of epistemically rewarding.
4. Finally, we outlined some choice points for that theory, using Evans' notion of a dominant source of information.

As in the case of predicates discussed in Chapter 5, this proposal is schematic, though not more so than the theories it is modelled on. It reinforces the conclusion from Chapter 4: the standard internalist approaches to AI have significant limitations. We cannot discover all the facts about the contents of the machine learning system's classifications simply by looking at the internal programming implementation of those systems. What discussion in this chapter concludes is that we need to focus in particular on the interpreters' epistemic goals and activities. What the AI system is about can vary depending on the interpreter's aims.

# APPLICATION

*Predication and Commitment*

In the previous two chapters, we've sketched metasemantic stories that specify the facts in virtue of which SmartCredit refers to Lucie and the property *high risk* in its output tokening of 'Lucie is high risk'. However, getting metasemantic stories that ground the meanings of these component parts of SmartCredit's output is arguably insufficient. We also need to explain why SmartCredit, in combining the expressions 'Lucie' and 'is high risk' in its output, doesn't just mention a person and a property, but also predicates the property of the person. In short: can SmartCredit (and other AI systems) produce or entertain or express full propositions?

In the philosophical literature, there is a very venerable literature about the nature of propositions. Our focus in what follows will not be about whether propositions are abstract entities, structured, etc. Our concern is primarily with the phenomenon called variously *entertaining* or *expressing* a proposition by predicating a property of an object. We are interested in predication qua an important semantic phenomenon, not qua a solution to the problem of the unity of propositions or the nature of propositions.

We'll pause here for a moment and recall the unrefuted sceptic from Chapter 2: according to him, what we are doing now is adding mistakes on top of our earlier mistakes. The earlier mistakes, according to the sceptic, were to look for genuine reference in the output of AI. We hope that a sceptic reading the previous chapters would react a bit concessively: maybe, they might say, there's a case to be made for reference to the property of being high risk and to Lucie, along the lines suggested. However, that need not be accompanied by conceding that the further step of finding a complete proposition (*that Lucie is high risk*) in the AI output is reasonable, nor agreeing that we should attribute to the system a commitment to that propositional content.

As before, our response is this: while we're not unsympathetic to the sceptic's position, we think the best way to test it is to try to see if an account of propositional content can be found. This is particularly *important* because there are hardly any efforts to do so. If you end up convinced that what we are about to propose is a potential way forward, then that's progress. If you find our effort entirely unconvincing, then it's a bit of additional support for the sceptic.

## Predication: Brief Introduction to the Act Theoretic View

To understand the full range of AI content, we need to think that SmartCredit can not only denote Lucie and the property *high risk*, but also *predicate* the property of being high risk to Lucie. In order to create a model of how that can happen, we need to understand the act of predication. Here we encounter the same kind of dilemma we've faced throughout this book: there are very many theories of predication. Our book is brief and we cannot comprehensively engage with that entire literature. What we will do instead is explore these issues against the background of a particular theory: the act theoretic view. An obvious drawback of this strategy is that if you, our reader, is adamantly opposed to this view, what follows will at best be of conditional interest to you. We hope, however, that the general strategy (of de-anthropocentrizing a view of predication) will be of use to you, even if the way we implement it isn't.

The view we will work with for the sake of argument is found most recently in the work of Scott Soames and Peter Hanks (we'll mostly be focusing on their respective monographs, both from 2015, but see also Soames 2019, and, for a slightly larger overview of the logical space of contemporary theories in this vein, Soames et al. 2014). According to Soames and Hanks, propositions are act types. They don't have intrinsic representational properties. Token acts (of the relevant type) are the original bearers of representation, truth conditions, and truth value. This is a change from traditional ways of thinking about propositions and representational objects more generally. They've traditionally been thought of as entities that somehow had an existence independent of the act of grasping them. The Fregean picture was of the mind as reaching out to—or grasping—entities that had existence independently of the act of grasping. The act type view reverses that picture: the primary explanatory element is the act of predication. Propositions understood as abstract objects, are act types instantiated by those token acts.

What is the act of predicating? Soames takes this to be a primitive notion. It picks out something we can easily recognize. For example, to think that *Lucie is high risk* involves denoting high risk, denoting Lucie, and then predicating the former of the latter. This token act of predicating belongs to a type. That type is the *proposition that Lucie is high risk.* The proposition that *Lucie is high risk* (i.e. the act *type*) has representational properties derivatively: it's the *token* act of that type that is true or false.

Is it an objection to Soames's view that he doesn't offer us an analysis of predication? He thinks not, and we agree. As he writes:

> One might ask what we mean by 'predication'—what, in effect, the analysis of predication is. Although it is unclear that an informative answer can be given to this question, it is equally unclear that this is anything to worry about. Some logical and semantic notions—like negation—are primitive. Since this elementary point typically doesn't provoke hand-wringing, it is hard to see why the primitiveness of predication should. (Soames 2015: 30)

Although primitivism about predication is defensible, it's worth considering at least one opposing view which tries to say more. The essence of predication, according to Hanks, is the ability to sort things into groups:

> Acts of predication are acts of sorting things into groups. When you predicate a property of an object you sort that object with other objects in virtue of their similarity with respect to the property. To predicate the property of being green of something is to sort that thing with other green things. This act of sorting can be done behaviorally, for example by picking the object up and putting it with other green things, or it can be done in thought, by mentally grouping the object with other green things, or in speech, by saying that it is green. (Hanks 2015: 64)

So understood, the *ability* to predicate, i.e. categorize, is a basic biological function that human beings share with the rest of the animal kingdom. Hanks approvingly cites Susan Gelman, who says:

[A]ll organisms form categories: even mealworms have category-based preferences, and higher-order animals such as pigeons or octopi can display quite sophisticated categorical judgments.

(Gelman 2003: 11)

Hanks uses the example of sniffer wasps (and bees) to illustrate how basic this ability is. Sniffer wasps can be trained to do various things, for example, detect landmines (and also various narcotic substances). In so doing, the wasp, according to Hanks, predicates the property of smelling like—for example—TNT to various objects. The act of predication, on this view, is the act of flying to those things.[1]

In the rest of this chapter, we explore whether a de-anthropocentrized version of the act theoretic view can be applied to AI systems. If that can be done, there is some evidence that the final step is achievable: Not only can SmartCredit denote the property of being high risk and Lucie, it can also predicate the former of the latter.

## Turning to AI and Disentangling Three Different Questions

In order to understand predication in AI, we'll disentangle three difference questions that are not always separated in the literature:

*Q1: What is it in a sentence that means predication?* There are several candidates for what that may be:

---

[1]  Absent the ability to use language to characterize its mental act, there is an irresolvable indeterminacy in the content of the wasp's judgement. Nothing fixes *what* the wasp judges or believes.

(i)  the concatenation of subject and predicate;

(ii)  the space between subject and predicate; and

(iii)  the root note in a tree having subject and predicate as daughter nodes.

The choice of syntactic object will depend on your background syntax and assumptions about how syntax, at the most basic level, interfaces with semantics. We won't take a stand on that here (for some helpful discussion, see King 2007 especially 33–6). What is important is that there's some syntactic feature that means or expresses predication. The question of how that syntactic feature ended up expressing predication is different from the question of what predication is, just as the question of how names—for example—ended up referring is different from the question of what reference is.

*Q2: Which mental act is the mental act of predication?* According to the act theoretic view, there is some kind of mental act that humans (and other animals, e.g. wasps) can perform. That act is, in part, constitutive of propositions. Since this act will be implemented in different ways in different animals, we can ask: how do we identify the particular mental state that is the act of predication? Moreover, since it's plausible that syntactic predication expressed in language somehow represents the mental act of predication we're assuming, having a grasp of what the latter is like will help us understand what syntactic predication expresses.

*Q3: The metasemantics of predication*: Having distinguished (1) and (2), we can distinguish two metasemantic questions:

(a) Why does this bit of syntax (i.e. the act of concatenation) mean predication? How did that part of syntax end up

having that meaning? (Analogous to how we can ask: how does a name end up having the referent that it does?)

(b) Why does this mental act mean predication? That might seem like a surprising question, but the underlying motivation for it is this: it is not just the intrinsic features of that act which makes it an act of predication. We'll see below that it is, in part, the functional role of that act— and we'll suggest that functional role can be spelled out in a teleofunctional way.

Note that both of these lattermost questions are questions in metasemantics that pattern with the metasemantic questions we have discussed about 'high risk' and 'Lucie' in Chapters 5 and 6. The three questions are closely connected. For Soames or Hanks, the answer to question (1) is that the relevant bit of syntax roughly expresses the performance of the mental act. The answer to (1) therefore partly depends on the answer to (2). However, the answer to (1) is not fully derivative of the answer to (2) because in order to answer (1), we need two things. Firstly, we need an answer to (3a), i.e. an account of why concatenation (or whatever) is hooked up to this mental act, rather than to something else. We also need an answer to (3b), i.e. a metasemantic explanation for why the mental act means what it does.

# The Metasemantics of Predication: A Teleofunctionalist Hypothesis

At this point we are going to add a new theory to the mix: teleofunctionalism. We do this for three reasons. First, it seems to us natural to think that this is, at least in part, a motivating thought behind the act theoretic view. Second, we want to use this as yet another illustration of how externalist theories can be helpful in giving an account of interpretable AI. Third, in de-anthropocentrizing the notion of predication we find in contemporary philosophy of language, we will need to find some way of identifying predication independently of its realization by human mental states, and functionalist theories in general are well-placed to do this.

As we interpret the act theoretic view, it identifies a mental act, $A$, that we perform. What makes $A$ an act of predication? Well, it's because the characteristic function of that act is to give rise to states of belief/judgement in us, which then give rise to characteristic kinds of behaviour. There's a familiar functionalist story sitting in the background here—to know what something represents, think about what kind of representational content would make sense of that something as a mediator between its characteristic inputs and outputs. It's then *teleo*functional because what matters is not the actual input/output performance, but what it's intended/designed/evolved to do. We can summarize this as the TF-Hypothesis:

> *TF-Hypothesis*: A mental act is the act of predication because of its teleofunctional role in giving rise to judgements that guide action.

The basic idea is that no mental act can be the act of predication in isolation from the function it performs. The relevant function is that of giving rise to judgements that then guide action. If a mental act doesn't give rise to judgements that play a role in guiding action, it would not be the act of predication. In the case of the sniffer wasp, the act of flying to a location is followed by acts of trying to extract sugar (that's how they are trained:

the TNT is laced with sugar and so the sniffer wasps are conditioned to fly to objects that smell like TNT). In the case of humans, the actions we perform are infinitely richer and impossible to make precise (and the claim we are making doesn't require that it can be).

In sum, the first part of the proposal is this: (i) a mental act cannot be the act of predication in isolation from function; (ii) the relevant function is that of giving rise to judgements that guide action; and (iii) that we perform acts with this function can be explained along teleofunctional lines. This gives us a way of identifying predication of the sort that Soames and Hanks are concerned with without committing ourselves to any particular architecture that implements it.

## Some Background: Teleosemantics and Teleofunctional Role

To understand our appeal to teleofunctional role, it will be helpful to say a few words about teleosemantic theories more generally. The basic thought is that the very fact that we (and, as we're arguing, AI systems) have content, as well as the particular contents we (and they) have is to be explained in terms of the idea of *function*. Paradigmatically, for teleosemanticists, these functions are biological. Thus, for example, the function of the heart is to pump blood; that's what it's there for, where in turn this notion of a particular thing being *there for* a particular function is to be cashed out in terms of evolutionary history—of natural selection. We evolved hearts because they are efficient ways to help spread oxygen and nutrients around the body (roughly).

The basic idea behind teleosemantics is that we treat representations just like we treat any biological adornment. Why does a cat represent birds? Well, why does a cat have whiskers? The answer to the latter question—roughly, we aren't vets—is that whiskers help cats navigate tight spaces. Having whiskers confers an advantage on cats, an advantage that in their evolutionary prehistory made them more apt at navigating their environment than similar felids which lacked whiskers. The whiskered cats successfully reproduced more, and so whiskers were selected for.

The same applies to contents. A cat that has no capacity to represent birds is a poor cat. It will miss out on opportunities for food, and so is less likely to thrive and reproduce, and so less likely to produce other cats. Cats that can represent birds will be well-fed and attractive mates, and more likely to produce other cats, which will be more likely than not themselves to be able to represent birds.

There are many teleosemantic theories, and many objections to teleosemantic theories, and while it's beyond both our aims here and our ability to decide between them, it will be useful to consider briefly some options and live questions which are relevant for this book. So, we might wonder what sort of representations we can attribute on the basis of teleosemantic reasoning. We considered attributing content like that associated with the full sentence 'there is a bird there' to our cat, but can we also attribute sub-sentential contents, such as representations of CATS, or—more abstractly and difficulty—BEING THERE? Theorists like David Papineau (presented in, for example, 1987, and more recently defending against some objections in 2001) think that our content attributions should be 'top-down', concentrating on representations of full beliefs and desires primarily as opposed to their component concepts (if the latter exist at all). Others—and this will arguably be particularly relevant for us—point out that this theory struggles with creatures that lack evolutionary history, or with creatures whose evolutionary history didn't involve the things they represent. It's good that we don't attribute iPad thoughts to cats, because they evidently don't care about them. We do greatly care about iPads, but iPads don't really figure in the history—on the evolutionary timescale—of our species. Davidson's famous Swampman example (1987) is of a creature molecule by molecule identical to us created by some—say—quantum mechanical fluke which, as such, has no evolutionary history (because it isn't the child of a child of a child…whose lineage is shaped by natural selection). To such a creature, it seems, we can't attribute any selected-for biological function, and thus no content—an arguably bad result.

Some wonder whether behaviour and evolutionary history are sufficient to give us *determinate* representations. Thus Fodor (1990) complains that considering functions will run into extensionality problems familiar from the philosophy of content in general. If we want to attribute to a frog the concept FLY because it behaves as if it has the concept, should not we equally want to attribute to it the concept of SMALL, DARK, FLYING THING? After all—at least if we stipulate that all and only flies are small, dark flying things—the two functions seem identical from a biological point of point. Content, the objection goes, is indeterminate in a way that is unattractive.

Let us emphasize: this is very much scratching the surface of a gigantic debate. There are many sophisticated accounts out there and an ongoing research program concerned with dealing with issues like the above.[2] But, for our purposes, it doesn't matter. Just as we were happy to take the basic Kripkean picture, deanthropocentrize a bit, and see how far we could get, so the foregoing superficial survey of teleosemantics suffices for our purposes.

We use some of the ideas behind basic teleosemantics in a new way: to give an account of how a certain state can mean predication and also to give an account of what predication is.

## Predication in AI

For AI, we can raise questions analogous to Q1–Q3. Start with the assumption that there is some aspect of the AI output that expresses predication. There is an initial question how to identify that external aspect of AI predication. Here are some options:

- If the output is linguistic in form, then maybe the answer here is the same as the answer in the normal linguistic case.
- However, not all AI outputs need be in linguistic form: If AlphaZero just moves pieces on the board (maybe it's connected to a magnetic system that lets it move the pieces), we can ask: what aspect of its output counts as the predication of 'good to move to A4' to the queen? One possible answer appeals to teleofunctionalism: that's the designed function of that aspect of the output.

---

[2] Thus we haven't mentioned Ruth Millikan's seminal work (1984, 1989a,b), or Nicholas Shea's sophisticated recent account (2018). And we haven't considered the important work of Karen Neander (2006, 1991, 1996). Again, the sole reason for this is that we don't think it has immediate bearing on the points to be made in this chapter. For an overview and more references, see Neander (2018).

For our purposes, we don't need to take a stand on the correct account of how predication is realized in the system. We'll just assume it is realized in some aspect, *A*, of the output. We can then ask (on analogy with Q1): *why* does *A* mean predication?

There is also an analogy with question Q2 above: we can assume that as in the human case, there is an aspect of the machine's 'inner life' that is predication (on analogy with the 'inner' human judgement that can be expressed, e.g. linguistically). Again, it's an open (and interesting question) what this is. Some options include:

- the system's computation;
- internal contentful states of the sort that get called 'intentional internals' in the AI literature; and
- some interactive aspect of how we use the AI.

If there is some such inner state, *ST*, then the feature we called *A* above only expresses predication derivatively: it is teleofunctionally connected to *ST*. *ST*, on this view, is the 'real' act of predication. *A* expresses predication derivatively, by being teleofunctionally connected to *ST*.

## AI Predication and Kinds of Teleology

Our proposal has incorporated an appeal to teleofunctionalism. One issue (of the many issues) we have not yet addressed is: *what kind of teleology are we talking about when we talk about teleofunctionalism?*

In answering this question, we can be guided by the meta-metasemantic principle in Chapter 4: knowledge-maximization. That principle guided our metasemantic theory (which in turn guided our interpretations). We can appeal to it here again in order to determine what the correct notion of teleology should be. The question then is: what kind of telos would be knowledge-maximizing for us interpreters, if we took it to be the teleofunctional role that makes a state into the state of predication?

The answer to this is far from obvious. Here are some options:

- The first place to look is at the design stage: the AI is designed by humans (or, sometimes, other AIs) and the designers will have in mind a functional role for the AI. So the simple answer is: the telos of the system is derived (maybe in some complex way) from that of its designers' intentions. It is that intended function by virtue of which some internal state (or derivatively, an external manifestation) is predication.
- Alternatively, we could treat AIs as more wasp-like. We would then ask what promotes the AI's own survival. 'Survival' in this case would need to be de-anthropocentrized—we look for whatever is the equivalent of survival for the AI system.
- A third alternative is to think about the goal, as derived from humans, not as derived from the AIs survival, but about the human-AI system as a whole. In this case, the telos is of the combination of humans and AI, not of one of those in isolation.

We are simply listing these as options. Our goal here is not to argue for a particular answer, but to show that there's a rich field of inquiry that opens up when teleofunctionalism is introduced to help us understand and interpret AI.

# Why Teleofunctionalism and Not Kripke or Evans?

A reader might reasonably say: in previous chapters we appealed first to Kripke (to explain AI denotation of high risk), then to Evans (to explain AI denotation of Lucie), and then now we are suddenly using a teleofunctionalist framework to explain predication. The reader might then ask: what is going on here? Aren't these competing frameworks? How can we selectively endorse all of them?

The first part of this answer to this is that we are not endorsing any of these metasemantic frameworks. Our goal is to show how they can be developed and adopted to understanding AI. We do so by de-anthropocentrizing guided by knowledge-maximization. If one or more of these strategies are promising, that's at least the beginning of a reply to the representational sceptic, who argues that this project is not even worth pursuing. It is also an argument in favour of exploring the various externalist traditions in the philosophy of language. That tradition has not been sufficiently exploited in this domain. In short, part of the answer to 'Why teleofunctionalism?' is: just so we can talk about another tool in the externalist toolkit, and continue to develop a general 'think about the externalist relations of the AI system, not its internal computational states' theme.

More specifically, it's hard to see how any kind of tracking/anchoring story could work for predication, because it's not clear what the thing to be tracked/linked is, or what it would mean to track or link to it. Predication is the classic syncategorematic item, where you want to give meaning not directly, but via how it affects the meanings of other stuff. Conceptual role semantics is a natural thing to use for syncategorematic items (that's why 'and' has always been the best case for conceptual role semantics), and conceptual role semantics is really just a special case of functionalism.

## Teleofunctional Role and Commitment (or Assertion)

So far, we have sketched an explanation of what it is about SmartCredit (internally, externally, or both) that makes it the case that SmartCredit has propositional content: that it doesn't just refer to Lucie and express the property of being high risk, but that it has the content that *Lucie is high risk*. So far, we have not explored the question of what it is about SmartCredit that makes it the case that SmartCredit *takes a stand* on that proposition: that it asserts that Lucie is high risk, or concludes that Lucie is high risk, or suggests that Jones is high risk. Here is Soames on the distinction:

> Although to entertain the proposition that o is red is to predicate redness of o, and so to represent o as red, it is not to commit oneself to o's being red. We often predicate a property of something without committing ourselves to its having the property, as when we imagine o as red, or merely visualize it as red. Hence, predication isn't inherently committing. Nevertheless, some instances of it, e.g. those involved in judging or believing, are either themselves committing, or essential to acts that are.    (Soames 2019: 2)

This is a point where Soames and Hanks disagree. As Hanks puts it:

> an act of predicating greenness of something is correct just in case that thing is green. Correctness and incorrectness here are just truth and falsity…This means that the act is true just in case that thing is green. An act of predicating a property of an object is true or false insofar as it can satisfy or fail to satisfy the correctness conditions determined by the property. Acts of predication have truth conditions and truth-values.    (Hanks 2015: 66)

For those sympathetic to Hanks's view, commitment/assertion is built into predication and so we wouldn't need a separate section on this.[3] For the sake of argument, we will tentatively assume Soames' view and see what can be added to get us from AI-predication to AI-commitment.

# Theories of Assertion and Commitment for Humans and AI

The question, then, is whether we can think of ML systems as committing to a content. To answer that, we need an account of what goes into that kind of commitment. Again, there's a massive literature on this as applied to humans (for which see e.g. Brown and Cappelen 2011, or Goldberg forthcoming).

To explore the issue of whether ML systems can perform speech acts, we could proceed as we did above: we look at various theories of what it takes to perform speech acts, and then see whether ML systems satisfy those conditions. If we focus on just assertion, there are at least four categories of views:

(i) Assertions are those sayings that are governed by certain norms—the norms of assertion.

(ii) Assertions are those sayings that have certain effects.

---

[3] But it raises the question of how to understand embedded propositions in negation, conditionals, etc. See Hanks 2015: ch. 4 for further discussion.

(iii)  Assertions are those sayings that have certain causes.

(iv)  Assertions are those sayings that are accompanied by certain commitments.

Within each of these categories of views, there's a great deal of variation. For example, there are very many norm-based views and no agreement about what the relevant norms are. Here are some of the more prominent suggestions (see Cappelen 2011: 9 for this taxonomy and references):

*Truth rule*: One must: assert p only if p is true.

*Warrant rule*: One must: assert p only if one has warrant to assert p.

*Knowledge rule*: One must: assert p only if one knows p.

*BK rule*: One must: assert p only if one believes that one knows p.

*RBK rule*: One must: assert p only if one rationally believes that one knows that p.

Other theories of assertion construe it as an act of commitment. This view is found in a range of authors, going back to Pierce and continuing with people such as Brandom (Pierce 1934; Searle 1969; Brandom 1994). Here is a version of the view from John MacFarlane:

(W*) In asserting that *p* at $C_1$, one commits oneself to withdrawing the assertion (in any future context $C_2$), if *p* is shown to be untrue relative to context of use $C_1$ and context of assessment $C_2$.

(MacFarlane 2005: 320)

Here is a research project: for each of these, explore whether these are norms that can be followed by an ML system. Despite their differences, they all raise the more general issue: what is it to follow or obey a norm and is that the kind of thing that an ML system can do? To investigate that question, we need an account both of the nature of norms and of what it is to follow them. Our prediction is that doing so will require using many of the same strategies we used above: you'll find current theories parochial because of being too anthropocentric. Then you will need to engage in anthropocentric abstraction, and you'll find some way to create a notion of 'assertion' or 'saying' that can fit ML systems. This will have the added effect of improving normative theories of assertion or saying.

In this book we will not carry out this project, in part because one of the authors of this book is sceptical of the very category of assertion (Cappelen 2011) and the other sympathetic to the Hanks' view that predication is committal (and so the theory of predication is all we need). That said, for those who want to pursue this project, the general meta-metasemantic principle from Chapter 4 should still be of help. Applying the knowledge-maximizing principle, we should expect the speech act of assertion to be such that it is knowledge-maximizing: we should expect assertion to be the kind of thing that maximizes knowledge for the audience member (i.e. the interpreters). If you were to pursue that line, the Williamsonian view that assertion is governed by the knowledge norm is tempting. However, endorsing that view also involves accepting that there are constitutive norms of assertion. That is an additional controversial assumption, but not one we will explore further in this book (but see the references above, in particular the anthologies and handbooks, for much recent work from many different perspectives).

# PART III
# CONCLUSION

# FOUR CONCLUDING THOUGHTS

This book does not aim to be a comprehensive treatment of all issues relevant to AI interpretability. That would require much more than what we have provided here. We have tried to focus on a small subset of issues that is relatively self-contained: how metasemantic work in the externalist tradition can be used to create models for AI interpretability. In this final chapter, we will end with four scattered thoughts that we hope will illuminate and develop some of the ideas in the previous chapters:

1. The first issue we address is an important direction for further work: philosophers need to engage in more detail with the fact that AI systems have certain kinds of dynamic goals.

2. Second, we explore what happens when someone sympathetic to the views in this book also endorses a version of Clark and Chalmers' thesis of the extended mind.

3. We revisit an objection to our entire approach: that we have not sufficiently explored the idea that we should give up talk of AI systems being representational or treat such talk as a form of make-belief.

4. Finally, we return to the topic of explainable AI from the introduction, indicating what we think can be learnt about that movement from the metasemantic perspective taken in this book.

# Dynamic Goals

The dynamic nature of neural networks gives rise to potential (and maybe actual) situations that makes the systems fundamentally uninterpretable. In the discussion above, we have conveniently ignored this feature of neural networks, but that's maybe cheating slightly. Below we give a brief outline of the sorts of issues that arise and require further investigation.

We start with a little story that illustrates what we have in mind. We should emphasize that this is not science fiction—it is, in effect, a partial feature of all neural networks. Our story is just meant to highlight something that we have not sufficiently focused on.

## A Story of Neural Networks Taking Over in Ways We Cannot Understand

Suppose you've decided to build a machine learning system to help a bank run its business. You start with a very specific mandate. The bank makes many loans to individuals—some of these loans are repaid, but some are defaulted on and not repaid. The bank wants to minimize the number of defaulted loans. The difficulty, of course, lies in spotting, among loan applications, the likely repayers and the likely defaulters.

Naturally enough, you begin with a supervised learning project. The bank gives you access to thousands of actual prior loans and their eventual outcomes (repaid or defaulted). You train a machine learning system by giving it details about each loan and testing your system's classification against the real loan outcome. (Of course, what details to give it about loans will be one of the difficult points. Perhaps you begin by giving the information provided on the loan application—income, savings, and some basic demographic information. But then you discover you get better outcomes by providing more input data. Eventually, following the pattern of companies like SmartCredit, you use the full social media history of a loan applicant as part of the input data for classification.) With adequate training, your program gets very good at sorting loan applications into defaulters and non-defaulters.

After a while, though, it occurs to you that you might do better. Your current program is very good, but not perfect, at finding defaulters. It makes occasional mistakes in both directions: sometimes it flags a loan application as a likely default when in fact the applicant doesn't default (a false positive), and sometimes it doesn't flag a loan application as a likely default but the applicant does in fact default (a false negative). Both false positives and false negatives are costly. False negatives directly cost the bank through loans that aren't repaid; false positives cost the bank by denying it access to potential interest income. If your program could be perfect, eliminating all false positives and all false negatives, that would be ideal. But that's not realistic—the available data just doesn't definitively and perfectly reliably settle the outcome of every loan. Mistakes are going to happen.

Your goal so far has been to minimize mistakes. But you realize that that might not be ideal. Some mistakes are much costlier than others. False negatives are costlier than false positives. And mistakes on large loans are costlier than mistakes on small loans. So you change the reinforcement learning pattern for your system. Now instead of just giving it a yes/no, default/no-default feedback on each loan application it evaluates, you give it a damage score feedback, telling it the amount of money that its evaluation has cost the bank, in light of the true outcome of the loan. The system is then trained to minimize damage scores.

It could well happen that the result is an increase in overall error rates in making default judgements. The machine learning system becomes sensitive to different patterns in the data, and those patterns aren't as well coordinated with the question of whether the loan will be defaulted, so it makes more mistakes of that sort. But the new patterns are well coordinated with something like *costly default*. One way to put this is that your program has changed from being a default detector to being a costly default detector. When we think of it in this way, there hasn't really been an increase in the error rate. It's true that more often now the machine learning system says 'yes' for a loan that goes on to be defaulted on, and 'no' to a loan for which there isn't any subsequent default. But those are only *errors* if we think that the program's 'yes' means 'yes, this loan is safe from default' and its 'no' means 'no, this loan isn't safe from default'. If the machine has changed, by virtue of the new reinforcement pattern, from being a default detector to being a costly default detector, then its 'yes' now means 'yes, this loan is safe from costly default', and it isn't, in fact, making more mistakes.

The revised bit of financial software is a hit—bank profits go up as costly defaults are avoided. Encouraged, you look for further such modifications in your financial detector. You have a few ideas—maybe you could train it to minimize some product of size-of-default and low-size-of-bank-financial-reserves, or maybe you could train it to minimize loss of potential interest earnings, so that false negatives are allowed to increase when interest rates go up. But you're a programmer, not a financial wizard—you worry that while you've got a few ideas about what loan features should be detected, you might be missing important features. (Or making horrible mistakes about what features to fixate on— maybe for subtle financial reasons you don't grasp, it would be a disaster to bias towards giving out more loans when interest rates are high.)

So you have another idea. Why not just use the overall financial state of the bank as the feedback mechanism for your program? Let it experiment with accepting and rejecting loan applications in various ways, and just let it know as it accepts and rejects how the bank is doing. That way, if increasing loans when interest rates are high is a good idea for overall bank health, the program can hopefully eventually get on to that pattern. But if that's not a good idea for overall bank health, the program will avoid looking for that pattern. No need for you to use your own defective financial understanding in picking a pattern for the program to detect.

After much training, the new system goes into effect, and it's a big success. Bank profit margins, when making loans following the advice of the new system, go up sharply. The bank CEO comes by to ask you about this great new piece of software, and asks you what the program is looking for when considering a loan application. This looks like a hard question to answer. You used to know what the program was looking for—originally it was looking for loan applications with a high probability of default, and then it was looking for loan applications that it would be costly to accept. But with your final revisions, there's some important sense in which you don't know any more what the program is looking for. Maybe you were right and it's good to accept more loan applications when interest rates are high, and thus maybe the program is now looking for some feature involving interest rates. But you can't tell easily—you'd have to look over thousands of loan recommendations by the program to see if that pattern does indeed emerge. And that's only one thing the program might be looking for; one that happened to occur to you. Who knows what other subtle patterns might be hidden in the program's decisions?

## Why This Story is Disturbing and Relevant

It now looks like you're in a somewhat disturbing situation, because in some important sense:

A. you no longer know what the program is looking for, and thus you don't know what the program means when it gives a positive or a negative verdict; and

B. control over what content the program is using, what category it is testing for, has been taken out of the hands of you, the programmer, and given over to the program.

We could try to avoid this conclusion by saying that the program is investigating some high-level abstract goal. You trained the final version of the program by giving it information about the overall financial health of the bank and then asking it to approve or reject loan applications depending on whether they improved that financial health. So maybe that's what gives the content of the program's verdicts. Maybe when the program says 'yes' to an application, it is characterizing that application as 'a loan that will improve the overall financial health of the bank' (rather than as 'a loan unlikely to be defaulted on'). More generally, can't we always just ask what you trained the program to do, by asking what kind of scoring mechanism you used for its decisions, and then just read off from that what the content of its decisions are? If that's right, the contents of the states of a program can change, but not in any mysterious way—they change only when *we* change how we evaluate program outputs.

But here are two worries about this 'high-level content' response.

First, it seems like it's missing something important to attribute only the high-level content to the program. Set aside software for a moment. Suppose the bank hires a (human) financial advisor, and asks him to figure out rules for which loan applications should and shouldn't grant in order to maximize the financial health of the bank. The financial advisor hides away in his office for a while studying volumes of data about old loan applications, and eventually declares himself ready and starts evaluating loan applications. Things go very well—the loan decisions the advisor makes are working out to the advantage of the bank. So we ask him what the method is—what feature of applications does he look for in determining which ones to accept?

If he tells us that he looks for loan applications that have the property 'will improve the financial health of the bank' (mentioning, perhaps, that that is after all exactly what we asked him to find), we will feel that he is holding out on us, and not telling us the property that he's *really* looking for. What we want to say is that he's looking for some unknown property P, and he's looking for that property *because* having that property contributes to the overarching goal of improving the bank's financial health. The overarching goal doesn't set the content of his rules—rather, it gives the reason for his rules having the content that they do (whatever that is).

Similarly with the software. To say that it's detecting the property 'improves the financial health of the bank' seems like it's confusing what property it's detecting with why it is detecting that property. If that's right, then we *don't* know what property the program is detecting, and can't directly control what property it's detecting.

A second worry is this: the high-level content approach depends on us at least knowing what the scoring mechanism is for the program. But maybe that doesn't always happen. Suppose the financial software is designed so that in addition to changing its sorting procedures, it can also change its scoring mechanism. So *we* don't tell it to start favouring detection categories that maximize the overall financial health of the bank—it changes its own scoring mechanism to start favouring those categories. Of course, if the program isn't going to behave randomly, its own changes in scoring mechanism need to be rule-governed in some way. So perhaps the programmers give the program a second-order scoring mechanism for evaluating how well its choices of scoring mechanisms are doing. In that case there's an even-higher-level content that we could ascribe to the system: the program is detecting objects as 'being things that maximize fit with respect to some criterion that maximizes achievement of goal G', where goal G is what we've encoded in the second-order scoring mechanism.

And, of course, we can ascend another level to a third-order scoring mechanism, which lets the program pick its own second-order mechanism for assessing its own choices of first-order scoring mechanisms for assessing its own choices of classifications, and then test that choice against our third-order criterion. Again, a very abstract higher-order characterization of the content of the machine verdicts can be given in this way, but any worries we already had about whether this abstract higher-order content is missing something important are only going to be made worse.

There's no limit to how many levels we can ascend. One limit point of this procedure has us switching over from a supervised learning software design to an unsupervised learning data mining design. The unsupervised learner starts with a kind of higher-order scoring rule, which just characterizes certain kinds of very abstract mathematical structure in the data as being good. It then looks for such structure in the data, and characterizes things in terms of that structure. Then it looks at that characterization and again looks for the desired kinds of structure in it. And so on, until, hopefully, some interesting large-scale patterns start to emerge. We might set such a data miner to work on a large history of loan applications and other financial information about the bank, and then try out using its classifications in making decisions about approving and rejecting loans. If things work out well using its classifications like this, we could then conclude that the machine is getting on to some feature worth attending to, without having any idea what that feature is, and thus without having any idea of what the program is telling us.

## Taking Stock and General Lessons

It would be great to know even in these cases of complex dynamic shifting of program contents what these programs are telling us. After all, we may be handing off control over large aspects of our lives to such systems. If we're going to be denying someone a loan to buy a house based on the output of some program, it would be nice if we could tell that person *something* about why their loan was denied, what it was about them that made them not loan-worthy. If we're going to begin an aggressive course of medical intervention on someone based on the output of some program, it would be nice if we could tell that person *something* about why that medical intervention was called for, what it was about them that was unwell or would be made better. In the limiting case, if we hand off control over judgements to machine learning systems with dynamically shifting goals that we can't understand, there may be no reason to expect that the things that we're told to do are things that we *ought* to do in any sense.

Dynamically shifting program contents, in short, give us special reasons for wanting a good story about what makes programs about the things they are about and a good story about how to find out what programs are about, but also special reasons for thinking that it may be particularly difficult to get the good stories that we want.

## The Extended Mind and AI Concept Possession

### *Background: The Extended Mind and Active Externalism*

In this book we have drawn heavily on the externalist tradition in metasemantics. It's a tradition that traces back to the work of Millikan, Kripke, Marcus, Putnam, and Burge. There is, however, another tradition that uses the term 'externalism'. In a brief but massively influential paper, Andy Clark and David Chalmers defend what they call 'active externalism'. They argue for the view that the environment, what is found beyond the skull/bone boundary, can drive cognitive processes. Their form of externalism is one in which 'the human organism is linked with an external entity in a two-way interaction, creating a *coupled* system that can be seen as a cognitive system in its own right' (1998: 8). The result of this is a view according to which various external devices (which can include AI systems) should, under certain circumstances, be seen not just as cognitive tools, but as integral parts of human cognitive processes. They apply this view not just to cognitive processing, but also to, for example, beliefs. If a device external to skull/bone 'contains' information and an agent is appropriately related to that external device, then that information can be one of the agent's belief. For example, on the assumption that Lucie's phone is appropriately related to her, and that the phone contains the information that Nora lives in Pokfulam, then Lucie believes that Nora lives in Pokfulam, even if that information is inaccessible to her without the help of her phone. The phone, on this view, is an extension of Nora's mind, on par with the synapses and whatever else is doing work inside Nora's skull and bone. Clark and Chalmers say that this kind of view enables us to 'see ourselves more truly as creatures of the world' (1998: 18). It is a corollary of the view that we should also see the self as extended beyond skull and bone. The external tools that are parts of Lucie's mind are parts of her—they are her in the same sense as her brain or ear is her.

Clark and Chalmers emphasise that the external device plays 'an active causal role', in the following sense: the system as a whole (what's inside skull/bone + the device + the relationship between the device and what's inside skull/bone) jointly influence actions. This is, in all relevant respects, similar to what cognition usually does: 'Our thesis is that this sort of coupled process counts equally well as a cognitive process, whether or not it is wholly in the head.'

### *The Extended Mind and Conceptual Competency*

The kinds of externalisms we have relied on earlier in this book do not directly engage with action in the way, e.g. external electronic devices can do on Clark and Chalmers' view. For example, the distal sources of Kripkean causal chains (the dubbings) are not causing a speaker or thinker to turn left rather than right as she is walking down the street. Information on the iPhone, on the other hand, could have that kind of active impact on an agent's action. Hence the term 'active externalism'. As in the earlier part of this book, we will conditionally endorse this form of active externalism.

There has been a great deal of discussion and development of the Extended Mind Thesis and we will simply bypass that discussion. We want to focus on one potential corollary of the view that, to our knowledge, has not been extensively explored. Can the extended mind also have extended conceptual capacities? More specifically: suppose, as we have argued, that AIs can have conceptual content and conceptual competency. That view, combined with the Extended Mind Thesis, has as a corollary that we get/inherit that conceptual competency from those AIs that are part of our minds.

### *From Experts Determining Meaning to Artificial Intelligences Determining Meaning*

There is a way into this that doesn't require appeal to the Extended Mind Thesis. Suppose you are sympathetic to Putnam or Burge's

style varieties of passive externalism. Putnam's slogan was 'Meanings ain't in the head'. That raises the question: where are they? The answer is either *nowhere* or *somewhere outside the head*. If we insist on something location-like, what we typically get is an appeal to *experts*. Experts, we are told, have the authority to determine the extension of e.g. predicates for natural kinds. In Burge's arthritis example, the community of medical experts have made it the case that the term 'arthritis' denotes ailments of the joints.

If you are on board with this view, and you are on board with our view that artificial systems could have contents, then the meanings could be located in artificial experts as well as human experts. This is a very natural move and it is independent of the endorsement of the Extended Mind Thesis. One source of objections to that view is that artificial agents don't have meanings or representational capacities. We have argued against that view. If you're on board with our arguments, the step from Putnam to AIs determining meanings isn't that radical.

## Some New Distinctions: Extended Mind Internalist versus Extended Mind Externalists

Here are two interpretations of the core part of Putnam's externalism that are typically not clearly distinguished:

P1:   meanings are not located inside the speakers skull/bone.
P2:   meanings are not 'in' the speaker's mind.

A simplistic assumption to the effect that the mind is 'inside' the skull/bone would equate P1 and P2. If you endorse the Extended Mind Thesis, you could deny P1 and endorse P2. More generally, a broader range of possible positions open up when thinking about meaning externalism. Here are some of those options:

- **Extended mind internalist**: meanings are located in (supervene upon) the extended mind.
- **Extended mind externalist**: meanings do not supervene on what's in the extended mind.
- **Skull/bone internalist**: meanings supervene on what's inside skull/bone.
- **Skull/bone externalist**: meaning do not supervene on what's inside skull/bone.

Note that a skull/bone externalist can endorse either extended mind internalism or extended mind internalism.

### Kripke, Putnam, and Burge as Extended Mind Internalists

Classical externalists like Putnam, Burge, and Kripke all seem like they would most naturally be classified as extended mind externalists. The reason for this is that the communicative chains that Kripke appeals to don't play the same kind of active role as the various kinds of external devices that Clark and Chalmers use as their paradigms (e.g. notebooks and phones that are used to guide behaviour on a regular basis). The experts who play a meaning-constitutive role for Putnam and Burge are similarly distal.

However, on further reflection, this is not at all obvious. A lot will depend on how the relationship to external devices is understood. On that point, Clark and Chalmers are extremely open-minded—more so than is typically recognized. When summarizing the relationship, R, that they suggest needs to obtain between an external device and an agent for that device to be part of the extended mind, they first mention four factors:

**Constancy**: '…the notebook is a constant in Otto's life—in cases where the information in the notebook would be relevant, he will rarely take action without consulting it.' (17)

**Ease of access**: '…the information in the notebook is directly available without difficulty.' (17)

**Automatic endorsement**: '…upon retrieving information from the notebook he automatically endorses it.' (17)

**Conscious endorsement in the past**: '…the information in the notebook has been consciously endorsed at some point in the past, and indeed is there as a consequence of this endorsement.' (17)

These four factors, however, are simply presented as salient generalizations of some features of the examples discussed in the paper and not given a theoretical justification. A full theory would need a justification for each of these, discussions of other options, and precisification. Clark and Chalmers are aware of this. Towards the end of the paper, their view becomes very liberal and open-ended. They say that what is part of the extended mind can be indeterminate. They say that being part of the extended mind might come in degrees: something can be a bit, but not fully, part of someone's mind. Finally, whether something is part of the extended mind could depend on context and in particular it could depend on the question under discussion. In a certain conversational setting, E might be part of A's extended mind, but in other conversational settings E might be excluded:

In intermediate cases, the question of whether a belief is present may be indeterminate, or the answer may depend on the varying standards that are at play in various contexts in which the question might be asked. (17)

In another passage they say that other people could, in certain context, for certain purposes, when certain questions are under discussion, be part of an agent's extended mind:

[T]he waiter at my favorite restaurant might act as a repository of my beliefs about my favorite meals (this might even be construed as a case of extended desire). In other cases, one's beliefs might be embodied in one's secretary, one's accountant, or one's collaborator. (17–18)

Putting aside Clark and Chalmers' view, it should be clear that the exact nature of the relation an external phenomenon needs to stand in to a person in order to be part of that person's extended mind is unsettled. It is unsettled not just in the sense that we know too little about it, and so haven't found the answer. It is also unsettled in that our concept of 'mind', 'belief', 'desire', 'memory', and so on are in flux. Those concepts will evolve in part with the way we interact and engage with technology. A full exploration of this would go very far beyond anything we can cover, but we end this section with a couple of conjecture/proposals:

1. For certain purposes having to do with attribution of conceptual competency, other people, e.g. experts, can be part of an agent's extended mind. Suppose the expert opinion is very easily available in a reliable way (say through a device for accessing information on the internet) and suppose the agent defers in various ways to those experts. This doesn't look too different from the waiter or accountant case.

2. Speakers who are part of Kripkean communicative chains are in constant causal contact with that chain and, according to Kripke, intend to refer to whatever was at the beginning of the chain. The connection to the communicative chains is constant, easy, automatic, and deferential. So, when questions of conceptual competence comes up, causal communicative chains can be part of our extended mind.

If these hypotheses are correct, then Kripke, Putnam, and Burge should be classified as internalists in our new sense, i.e. they are extended mind internalists. Of course, we have made them internalists, by radically extending our notion of the internal. Rather than see the content-determining factors as factors outside the mind-determining content, we have extended the mind to include the content-determining factors.

### Concept Possession, Functionalism, and Ways of Life

Here is a natural thought: the kinds of concepts we have are related to the kinds of creatures we are and our way of life. Our conceptual repertoire is, in part, determined by us being certain kinds of animals, with certain kinds of inputs (in large part determined by our perceptual capacities), and certain kinds of outputs (our actions often involve movements of our physical bodies). Functionalists pick up on this basic idea and tie contents to the kinds of inputs and outputs that are possible for us. These functional roles, that are meaning determining, are fixed by the kinds of creatures we are and our way of life.

However, if extended internalism is true, then this is less of a limitation: we start out as certain kinds of animal with certain input and output capacities. Then we extend ourselves using, for example, artificial intelligence. This extension means that the range of contents we can entertain is extended because our possible input and output functions have been extended. This is because what we are has changed and our way of life has changed, as a result of ourselves being extended.

## Implications for the View Defended in This Book

The strategy in this book has been to start with anthropocentric views in metasemantics, do some de-anthropocentrizing, and then try to apply the result to artificial intelligences. The Extended Mind Thesis suggests a complimentary strategy: incorporate! We have been assuming that whatever the AIs are doing isn't what we humans are doing and so we need to find some common process at a higher level of abstraction (the abstractive sweet spot). The alternative strategy just explored thinks of those AIs as potential (at least to some degree and in some contexts) parts of our minds. If those AIs are part of our minds (or rather: their processes are cognitive processes on par with what's happening inside skull/ bone), then they are part of us (in the extended sense of us) and so what they are doing is what we are doing.

The effort to understand alien content determination thus becomes an effort to understand our own extended mind's content determination. That is a very useful perspective from which to approach these issues, but the issues that need resolution will be roughly the same as those discussed earlier in this book. Our aim will be to get a grip on how hard to understand extended parts of us (e.g. the artificially created neural network that, to some degree and in some contexts, are parts of our extended mind) determine content. Seen in that light, the project pursued in this book is an exercise in extended-self-examination.

# An Objection Revisited

We return briefly to an important thought that was behind some of Alfred's objections in Chapter 2. He was resisting the idea that questions about content had significance for his work in AI. We managed to persuade Alfred to take an interest in some of the philosophical issues we have outlined above, but of course, the way we wrote and ended that dialogue was self-serving: we gave ourselves what we needed. In some sense, we didn't do Alfred (or Alfred's position) justice. In particular, we didn't help him articulate an alternative to the content-focused picture that we have been pushing throughout. The objection we will now briefly address is this: Alfred should have focused on the notion of evidence or reliability—that's the alternative to a content-driven approach.

Here's a way to articulate that alternative:

> *The No-Content-Just-Evidence view:* Once StopSignDetector has been thoroughly trained on an initial sample of photographs pre-labelled as stop signs or not, we should then take the output of StopSignDetector as *evidence* that something is a stop sign, without thinking of StopSignDetector as having outputs whose *content* involves stop signs. StopSignDetector is, if nothing else, a *reliable detector* of stop signs. It is reasonable for us to form beliefs on the basis of the outputs of reliable detectors. Reliable systems can serve
as a form of evidence.

This view should be explored. It's an interesting alternative to the
strategy defended in this book. So far, our aim has not been on
developing direct objections to the No-Content-Just-Evidence
view, but rather to make an indirect case against it by developing
various pro-content alternatives. Those advocating for the No-
Content-Just-Evidence should do the same: develop positive
models that integrate theory of evidence and reliability with the
nature and use of AI. Note that this is again a fundamentally philo-
sophical project: it places philosophy at the centre of an under-
standing of AI. Those trained in computer programming, for
example, are not trained to think about the nature of evidence,
reliability, and how to apply theories of these phenomena to the
output and use of AI. A defence of No-Content-Just-Evidence view
would involve a shift in focus from the metaphysics of content to
the theory of evidence and reliability. For some initial literature on
this, see Kelly (2014) and references therein.

### *Reply to the Objection*

While we welcome an exploration of the No-Content-Just-Evidence
view, we are sceptical. We think it faces some serious obstacles, and
in the next couple of pages we briefly outline some of these.

### *What Makes it a Stop Sign Detector?*

The Evidence view, as we articulated it above, assumed that the
system in question was a reliable stop sign detector. It is not,

however, clear that we are entitled to that assumption. In some important sense, we have no idea what the mechanism is by which StopSignDetector reacts as it does. All we really know is that StopSignDetector has some unbelievably complicated neural network, with connections and connection weights developed over millions of rounds of training, that somehow or other filter through the incoming data from a photograph (initially presented in some data form or other—an array of pixel values, for example) to work out activation levels culminating in the light blinking or not.

The best we can say is that there is some structural property or other of photographs that StopSignDetector is *really detecting*. That structural property presumably is some enormously complicated property about hugely computationally demanding relations among many different aspects of the incoming numerically given data—almost certainly a property that no human mind could ever really grasp, and possibly a property that there isn't even a way to express in a human language. Call that structural property S. (Now there's a way to express it in our language!) If the existence of S is helping us decide that StopSignDetector is genuinely disposed to react to stop sign pictures, then we must have some reason to think that S and stop signs are reliably correlated—that, in general, when we get a new photograph of a stop sign, it's probably going to have property S.

But why would we think that? The training of StopSignDetector doesn't look like it gives us a good reason to accept it. Here's what we learn from the training. There are two big piles of test cases that StopSignDetector was trained on—call them the positive cases and the negative cases. Property S, whatever it is, must then be a property that most of the positive cases have and most of the negative cases don't have.[1] What reason, then, do we have to think that the particular property S that StopSignDetector got hooked onto is reliably correlated with stop signs?

## Adversarial Perturbations

We don't need to rely on abstract theoretical considerations like these. There is a growing body of work on adversarial perturbations (see, for example, Goodfellow et al 2014) and their impact on machine learning image recognition systems. Adversarial perturbations provide methods of making small alterations in images that result in a machine learning system going from a very high success rate in classification to a very low success rate. Adversarial perturbations can involve small alterations in the photograph that don't significantly impact human identification abilities—adding a few bits of coloured tape here and there to the object to be identified, for example, or slightly rotating the angle of photograph. Or they can involve adding a masking layer of pixels over the original photograph that the human eye can't even detect, but that causes massive misclassification by the machine learning system.

Note that to say that adversarial perturbations cause *misclassification* isn't really right, in the current context. It's a misclassification only if the system was really (for example) a stop sign detector, so that the adversarial perturbation is causing the system to get things *wrong* by saying that things that really are stop signs aren't stop signs. But that way of putting things is loaded up with content-based talk about what the machine really detects and says, and the current dispositional line is meant to be a replacement for that talk. From the current perspective, what's going on is that adversarial perturbations are revealing what structural property S the machine is really tracking.

---

[1] If there were any doubt about this, note that two different machine learning systems can be trained on the same data, using neural network systems with a stochastic element, and go on to make slightly different distinctions among new cases. That shows that they are really detecting different underlying structural properties. And not necessarily slightly different such properties—the detected structural properties could be radically different from one another, but have only slightly different distributions among the cases we've tested so far.

The worry, then, is that if StopSignDetector can be easily made to blink for non-stop-signs, or not to blink for stop signs, through using some adversarial perturbation that a human classifier would never even notice, then it's not clear that StopSignDetector is really disposed to blink when presented with a stop sign. The adversarial perturbation brings out the possibility of property S and the property of being a stop sign coming apart.

The general lesson: lots of things are statistically correlated with lots of other things. Dispositions require more than that. Dispositions require that the correlations be *reliable*, so that new cases will continue the correlation. When there's some underlying causal structure that created the correlation, we have a reason to think that it's reliable, even if we don't fully understand how the causal connection works. But in the machine learning case, we don't have any reason to think that there is a causal connection to support the disposition. That's because we do have reason to think that there's a different causal connection (one we don't fully understand) between some obscure structural property S and the machine learning outputs. Because we know there's that causal connection, it trumps the possibility of the causal connection we really want (but aren't getting). So we have to fall back on the coincidental convergence of weird structural properties and our target properties on the training cases—but the coincidental convergence doesn't give us reason to treat the system as reliable for new cases.

This is in no way a conclusive argument to the effect that the No-Evidence view could work. It is, however, conclusive evidence that doing so is far from trivial. It requires deep engagement with philosophy. The final view will rely on a theory of what dispositions are, what reliability is, and the connection between evidence and reliability. We'll end this brief reply with a suggestive conjecture: When you have a satisfactory theory of that kind—one that responds to all these concerns—you have in effect come very close to constructing a theory of content again. According to this conjecture, the Evidence and the Content strategies will merge.

## Explainable AI and Metasemantics

In the first chapter, we connected the topics of this book to the issues that come up in connection with the aim of achieving so-called explainable AI. 'Explainable AI' indicates a desire to ensure that decisions and other kinds of input made by AIs are not just handed down to us as from an oracle. If an AI system tells us that Lucie should not get a mortgage, she is entitled to understand why she should not get a mortgage. To answer the why-question by simply insisting that the decision was made by a reliable but incomprehensible algorithm isn't good enough. Lucie should also be able to understand why without having to perform the inhuman task of working through all the calculations made by an extremely complex neural network. She is entitled to receive a justifying reason for the rejection.

Even if everything we have said so far in this book is along the right track, we haven't succeeded in giving Lucie a procedure for getting a justification from SmartCredit. If we have succeeded, we have shown how SmartCredit can say that *Lucie is high risk*. That leaves us far short of getting a justification for *why* she is high risk. We have revealed nothing of the internal reasoning that might have gone into producing that output.

So did we engage in false advertising when we raised the prospect of illuminating explainable AI? Not really. We didn't claim we were going to show how explainable AI is possible. We did claim that our work could contribute to an understanding of how explainable AI could be possible. Here is how we see the connection:

1. Without content, there are no reasons. Reasons are things with content.[2] The natural way to think about the

---

[2] Actually, this is a bit controversial. While this book is not the place to get to grip with the vast and ever increasing literature on reasons, we'd just like to make clear that the above is a vast simplification that elides many important distinctions which any proper treatment of the topic will need to make room for. For example, a much-discussed question in the theory of reasons is taxonomic: how many different types of reason are there? To take an example from Alvarez's (2016) overview on the topic, reducing child obesity might be a reason for the government to tax sugary drinks in one sense, but a perfectly fine answer as to for what reason the government *in fact* taxed drinks in the winter of 2019 is because after the election the legislative body became filled with people who owned shares in bottled water companies. Roughly, the former would be a normative reason (something counting in favour of something in the abstract) while the latter a motivating one (something that in fact brought about a particular course in action). For a bit more on the distinction, see the opening pages of Dancy (2000).

Equally important are questions about the ontology of reasons. We've assumed they are items with representational content. It's the subject of further work exactly how that will work in the AI setting, because at least some of the literature has it that some reasons (normative ones) are non-representational entities like facts (Raz 1975 and Scanlon 1998). Moreover, a popular view about motivating reasons is that they are mental states (e.g. Audi 2001 and Mele 2003), and though this doesn't immediately cause a problem for us, there might be explainability desideratum is this: the AI says something—
e.g. that Lucie is high risk. Then Lucie is entitled to a
justification of that claim. For that entitlement to make
sense, the system must have said something (namely that
Lucie is high risk). If there's no saying, there's nothing to
justify. Moreover, the reasons themselves are contentful. So
we need content both to have something to justify and in
order to have something that can do the justifying. Here is
what we have done: we have provided a strategy for estab-
lishing that the system can perform sayings. In so doing,
we have shown how to take the first step towards
explainability.

2. Explainability requires not just the generic possibility of
content attribution, but also a procedure for determining
specific contents. We need to know exactly what the
system said before we can ask it to give a reason for what it
said. The most central claim in this book is that such
content cannot be found by looking at the internal
computational structure of the system. It can only be
found by looking at external factors of the kinds that the
externalist tradition in metasemantics appeals to. It is hard
to overemphasize this point. The story about AI content is
not substantially different from the human story: in neither
case do we find content by looking at internal computational
architecture.

---

questions to be asked about how AI, presumably lacking mental states, can be
present in the space of reasons. As mentioned, here isn't the place (and we
aren't the authors) to decide these issues. We are just flagging some important
distinctions an acceptable philosophical treatment of explainability will need
to grapple with.

We end with a brief explanation of what an account of reason giving (and justification) for a decision (or output) would require. First a reminder of some issues that would have to be resolved in order to present such a theory.

- The philosophical tradition distinguishes between at least three kinds of reasons for actions: normative reasons, motivating reasons, and explanatory reasons.[3] How to characterize each of these is a matter of ongoing dispute. So a first question to be settled is whether it is any of these is what we are looking for. Should we, for example, be modelling AI explainability on explanatory or motivating reasons in humans? If the answer to either question is yes, then a theory of explainable AI could incorporate an existing theory of motivating or explanatory reasons.
- Alternatively, a theory of AI explainability could develop a new such theory, maybe by engaging in some form of de-anthropocentrizing, on analogy with what we have done for metasemantics in this book.

---

[3] Often motivating and explanatory reasons are often classified together. We are sympathetic to those who prefer to keep them apart. Maria Alvarez in the SEP entry 'Reasons for Actions' nicely summarizes one argument for the distinction: 'The fact that John knows that Peter has betrayed him is a reason that explains John's action. This is an explanatory reason. But that fact about John's mental state of knowledge is not the reason for which John punches Peter. That reason is a fact about Peter, namely that he has betrayed John. That is the reason that motivates John to punch Peter—his motivating reason. So in this case we have two different (though related) reasons: that Peter has betrayed John and that John knows that Peter has betrayed him, which play different roles. One reason motivates John to punch Peter (the betrayal); and the other explains why he does it (the knowledge of the betrayal)' (Alvarez 2016: section 3).

Back to the human case: we humans have asked for and provided reasons (and justifications) for a very long time. It is an activity that is at the core of how humans relate to each other. That is in part why it is something we care about in connection with AIs. Here is a basic fact about the cluster of activities that we call 'giving reasons' or 'explaining' or 'justifying':

> **Important and indisputable fact about human explainability**: Humans have been able to explain and justify their own (and others) actions/decisions without relying on any knowledge of the internal structure of the neural network that constitutes (part of) their brains. Human explainability succeeds in the absence of any knowledge about the internal computational structure of the human brain.

What this tells us is that knowledge of internal computational structure is unnecessary for explainability. It is, however, exceedingly tempting to conclude also more broadly that such knowledge is irrelevant to explainability. If so, it's a mistake to approach the goal of explainable AI by careful investigation into the computational structure of the AI's neural network. That kind of internalism is bound to fail and it will never lead to the space of reasons.

# INDEX